\pdfoutput=1

\documentclass[11pt]{article}

\usepackage[preprint]{acl}

\usepackage{times}
\usepackage{latexsym}

\usepackage[T1]{fontenc}

\usepackage[utf8]{inputenc}

\usepackage{microtype}

\usepackage{inconsolata}

\usepackage{graphicx}
\usepackage{times}
\usepackage{listings}
\usepackage{multirow}
\usepackage{soul} 
\usepackage{subcaption}
\usepackage{graphics}
\usepackage{xspace}
\usepackage{amsmath}
\usepackage{amssymb}  
\usepackage{adjustbox}
\usepackage{makecell}
\usepackage{pifont}
\usepackage{tabularx}
\usepackage{xcolor}
\usepackage{booktabs}
\usepackage{float}
\usepackage{array}
\usepackage[T1]{fontenc}

\newcommand{\ourdata}{{\includegraphics[height=1em]{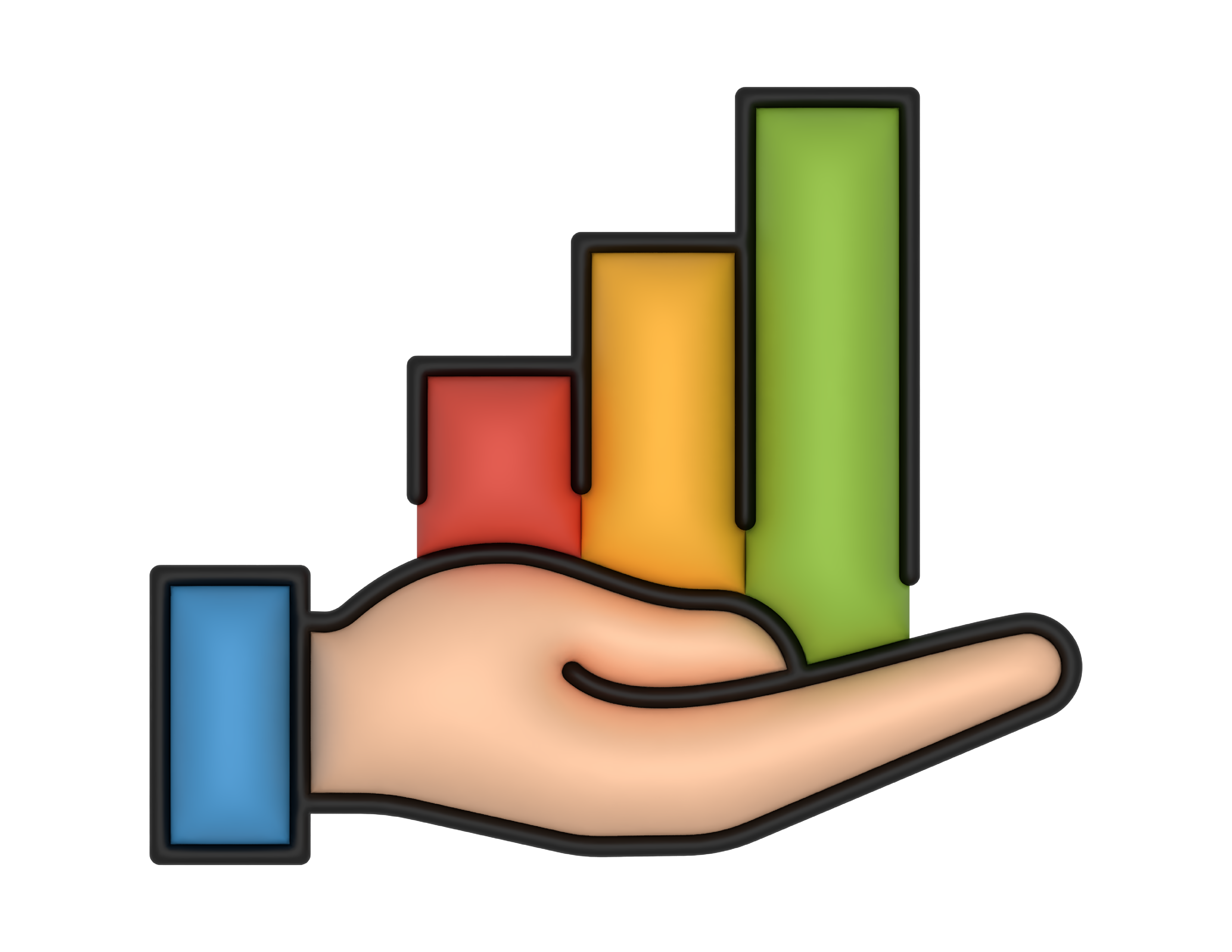}\textsc{MuSciClaims}}\xspace}
\newcommand{\ourdatasmall}{{\textsc{MuSciClaims}}\xspace}

\newcommand{\llm}{\textsc{LLM}\xspace}
\newcommand{\llms}{\textsc{LLM}s\xspace}
\newcommand{\vlm}{\textsc{VLM}\xspace}
\newcommand{\vlms}{\textsc{VLM}s\xspace}

\newcommand{\gfomni}{\texttt{4o}\xspace}
\newcommand{\gfmini}{\texttt{4o-mini}\xspace}
\newcommand{\sonnet}{\texttt{Sonnet}\xspace}
\newcommand{\othree}{\texttt{o3}\xspace}
\newcommand{\ofour}{\texttt{o4-mini}\xspace}

\newcommand{\phimm}{\texttt{Phi-4}\xspace}
\newcommand{\llava}{\texttt{Llava-Next}\xspace}
\newcommand{\llama}{\texttt{Llama-3.2}\xspace}
\newcommand{\molmo}{\texttt{Molmo}\xspace}
\newcommand{\ivl}{\texttt{InternVL3}\xspace}
\newcommand{\qwen}{\texttt{Qwen2.5}\xspace}
\newcommand{\ds}{\texttt{DeepSeek}\xspace}

\newcommand{\support}{\textsc{Support}\xspace}
\newcommand{\contradict}{\textsc{Contradict}\xspace}
\newcommand{\neutral}{\textsc{Neutral}\xspace}
\newcommand{\nsup}{\textsc{NonSupport}\xspace}

\newcommand{\phy}{\textsc{Physics}\xspace}
\newcommand{\chem}{\textsc{Chemistry}\xspace}
\newcommand{\bio}{\textsc{Biology}\xspace}

\newcommand{\overall}{\textsc{Overall}\xspace}

\newcommand{\dexpt}{\textsc{D}\xspace}
\newcommand{\rdexpt}{\textsc{R}$\rightarrow$\textsc{D}\xspace}
\newcommand{\irdexpt}{\textsc{I}$\rightarrow$\textsc{R}$\rightarrow$\textsc{D}\xspace}

\newcommand{\ver}{\textsc{ClaimVerification}\xspace}
\newcommand{\loc}{\textsc{EvidenceLocalization}\xspace}
\newcommand{\cma}{\textsc{Cross-ModalAggregation}\xspace}
\newcommand{\bv}{\textsc{BasicVisualUnderstanding}\xspace}
\newcommand{\rob}{\textsc{EpistemicSensitivity}\xspace}

\newcommand{\squishlist}{
  \begin{list}{$\bullet$}
    { \setlength{\itemsep}{0pt}      \setlength{\parsep}{3pt}
      \setlength{\topsep}{3pt}       \setlength{\partopsep}{0pt}
      \setlength{\leftmargin}{1.5em} \setlength{\labelwidth}{1em}
      \setlength{\labelsep}{0.5em} } }
\newcommand{\reallysquishlist}{
  \begin{list}{$\bullet$}
    { \setlength{\itemsep}{0pt}    \setlength{\parsep}{0pt}
      \setlength{\topsep}{0pt}     \setlength{\partopsep}{0pt}
      \setlength{\leftmargin}{0.2em} \setlength{\labelwidth}{0.2em}
      \setlength{\labelsep}{0.2em} } }

 \newcommand{\squishend}{
     \end{list} 
 }

 \newcommand{\xmark}{$\times$}

\title{\ourdata : Multimodal Scientific Claim Verification}

\author{
    \textbf{Yash Kumar Lal}\quad\quad
    \textbf{Manikanta Bandham} \quad\quad
    \textbf{Mohammad Saqib Hasan} \\
    \textbf{Apoorva Kashi} \quad\quad
    \textbf{Mahnaz Koupaee} \quad\quad
    \textbf{Niranjan Balasubramanian} \\
    Stony Brook University \\
    \texttt{ylal@cs.stonybrook.edu}
}

\begin{document}
\maketitle
\begin{abstract}
Assessing scientific claims requires identifying, extracting, and reasoning with multimodal data expressed in information-rich figures in scientific literature. 
Despite the large body of work in scientific QA, figure captioning, and other  multimodal reasoning tasks over chart-based data, there are no readily usable multimodal benchmarks that directly test claim verification abilities. 
To remedy this gap, we introduce a new benchmark \ourdata accompanied by diagnostics tasks. 
We automatically extract supported claims from scientific articles, which we manually perturb to produce contradicted claims. 
The perturbations are designed to test for a specific set of claim verification capabilities. 
We also introduce a suite of diagnostic tasks that help understand model failures.
Our results show most vision-language models are poor ($\sim$0.3-0.5 F1), with even the best model only achieving 0.72 F1.
They are also biased towards judging claims as supported, likely misunderstanding nuanced perturbations within the claims.
Our diagnostics show models are bad at localizing correct evidence within figures, struggle with aggregating information across modalities, and often fail to understand basic components of the figure.
\end{abstract}

\section{Introduction}

Scientific claim verification aims to assess the validity and correctness of a claim with respect to given scientific literature~\cite{kotonya-toni-2020-explainable-automated, saakyan-etal-2021-covid, mohr-etal-2022-covert,wadden-etal-2020-fact}.
Existing work on scientific claim verification mainly focuses on textual data. 
They pose verification tasks over a single article or text snippet \cite{kotonya-toni-2020-explainable-automated, saakyan-etal-2021-covid, mohr-etal-2022-covert}, a corpus of full-text articles \cite{wadden-etal-2020-fact}, or larger collections of scientific abstracts \cite{wadden-etal-2022-scifact}.

However, scientific evidence is often presented as heterogeneous information-rich figures that support the important findings, claims and conclusions of experiments. Therefore, scientific claim verification requires both textual and visual understanding capabilities. 
To assess a claim, one has to go over the figure and its caption, find the panel(s) with information relevant to the claim, combine this visual knowledge with textual information in the figure caption, and finally judging whether the claim is supported or not. 
While there is a large number of benchmarks on scientific figures, they focus on image captioning \cite{hsu2021scicap}, question answering \cite{kahou2017figureqa}, or other reasoning tasks \cite{yue2023mmmu}.
There are no readily usable multimodal benchmarks for scientific claim verification. 
The closest work, ChartCheck \cite{akhtar-etal-2024-chartcheck}, poses a multimodal claim verification task but is restricted to simple data charts crawled from the web, which are substantially different from complex figures found in scientific articles. 

To address this gap, we introduce \ourdatasmall\footnote{Data is available at \url{https://huggingface.co/datasets/StonyBrookNLP/MuSciClaims/}}, a multimodal benchmark for claim verification over figures in scientific (physics, chemistry and biology) literature. 
We set two desiderata for our benchmark: the dataset needs carefully constructed claims that are not supported or have contradictory information in the figures; 
apart from quantifying model performance, the dataset should also be diagnostic in nature to identify specific model weaknesses.
Our dataset creation methodology is designed to meet these desiderata.

We extract claims with inline references to figures from the results section of articles. 
We manually filter these to only retain claims that are clearly and unambiguously supported by the figures. 
Then, we create contradictory claims by perturbing these supporting claims. 
We devise a diverse set of perturbations to test specific capabilities for claim verification including qualitative and quantitative reasoning, and observation-inference connections.

Last, we create a suite of diagnostic tasks associated with each claim to better understand model failures. 
Specifically, we design tasks that help uncover errors across aspects of basic visual understanding, evidence localization, cross-modal aggregation, and epistemic sensitivity. 
We ensure the integrity of the dataset through manual analysis. 
The resulting dataset consists of 1515 \texttt{(claim, figure)} data points from \phy, \chem and \bio, equally balanced across 3 class labels (\support, \neutral, \contradict), each accompanied by diagnostic questions. 

We benchmark a suite of visual language models (\vlms) on \ourdata.
Most models are poor at scientific claim verification out-of-the-box.
Prompting \vlms to explain their decisions helps performance, but only slightly.
Despite these gains, there is still a large room for improvement.
Our diagnostics shows that models fail at evidence localization, introducing noise in their reasoning process consequently performing worse.
Their basic visual understanding and cross-modal aggregation capabilities also need improvement.

In summary, our contributions are:
\begin{enumerate}
    \item We present \ourdata, an evaluation benchmark for multimodal scientific claim verification over information-rich figures.
    \item We find that contemporary models are good, but have significant room for improvement on claim verification.
    \item Our diagnostic tests pinpoint specific model abilities to improve---localizing to the right information and cross-modal information aggregation---for better claim verification.
\end{enumerate}

\section{Related Work}

\paragraph{Multimodal Scientific Benchmarks}
There has been extensive work on evaluating multimodal understanding abilities of contemporary models.
Some work focuses on image captioning tasks where, given an image, the model is asked to generate a concise description for it \cite{hsu2021scicap,tang2023vistext}.
But the larger share belongs to question answering benchmarks. 
These benchmarks differ on types of image, questions, knowledge required to answer questions, domains, scale, and annotations.
While FigureQA \cite{kahou2017figureqa}, DVQA \cite{kafle2018dvqa} and PlotQA \cite{methani2020plotqa} provide large-scale resources, they are limited to synthesized charts and template-based questions. 
They do not fully capture the complexity and diversity of real-world charts.

To create more complex QA benchmarks, ChartQA \cite{masry2022chartqa} mixes 30k human and machine-generated questions; however the images are still limited to line, bar and pie charts.
To cover more types, ArXivQA \cite{li-etal-2024-multimodal-arxiv} extracts images with \llm-generated QA pairs from arXiv papers.
SciGraphQA \cite{li2023scigraphqalargescalesyntheticmultiturn} extract graphs from Comp. Sci. ArXiv papers and use \llms to create multi-turn question-answering dialogues about them.
MMC \cite{liu2024mmc} supports diverse tasks and chart types using free-form questions and open-ended answers. 

Previous benchmarks rely heavily on chart annotations or table metadata as textual prompts to generate content, allowing models to easily obtain candidate answers while ignoring the charts’ visual logic. 
ChartBench \cite{xu2023chartbench} includes both annotated and unannotated charts.
While ChartX \cite{xia2024chartx} covers more chart types, its data and charts are synthesized and limited to ones that can be directly converted into a structural data format, e.g., CSV format.
CharXiv \cite{wang2024charxiv} consists of 2k real-world charts with manually curated questions by human experts and answers validated by hand, panning 8 major subjects published on arXiv. 
The questions are either descriptive to understand basic chart data or reasoning-based to dig deeper into charts. 
MultiChartQA \cite{zhu-etal-2025-multichartqa} is designed to evaluate \vlms' reasoning capabilities across multiple charts.
However, the charts are not information-rich and no domain knowledge beyond what is stated in the charts is required to answer questions in these benchmarks.

To cover more image types, MMMU \cite{yue2023mmmu, yue2024mmmu} collected multimodal questions from college exams, quizzes, and textbooks, covering six disciplines, ranging from visual scenes like photos and paintings to diagrams and tables, testing the perceptual capabilities of \vlms.
CURIE \cite{cui2025curie} covers diverse scientific disciplines, but its multimodal tasks are limited to biodiversity georeferencing and protein sequence reconstruction tasks.
While this requires domain knowledge, it doesn't require a high level of expertise.

Most existing multimodal benchmarks are designed such that reasoning over just the images can result in the answer. 
But SPIQA \cite{pramanick2024spiqa} is designed such that questions require simultaneous reasoning over different modalities, including figures, tables, and texts from articles in the computer science domain.
Even with the diversity of multimodal benchmarks for scientific literature, there is a dearth of datasets to test how well models can verify claims made in such data.

\paragraph{Scientific Claim Verification}
Claim verification as the task of establishing the truthfulness of a given claim has gained a lot of attention given ever-increasing amounts of data \cite{thorne-etal-2018-fever, kotonya2020explainable, wadden-etal-2020-fact}.
Scientific claim verification requires significant domain knowledge as well as understanding the evidence to reason about the claim.
SciFact-Open \cite{wadden2022scifact} expand on previous work \cite{wadden-etal-2020-fact} to provide a more realistic testbed of claim verification systems.
Recent work has also focused on testing how well models can verify claims over tabular data on real-world public health claims and scientific papers \cite{akhtar-etal-2022-pubhealthtab, wang-etal-2021-semeval} or charts and images \cite{akhtar-etal-2024-chartcheck} or a mix of all \cite{singh-etal-2024-scidqa}.
\citet{akhtar-etal-2023-reading, akhtar-etal-2024-chartcheck} focus on claim verification over plots and charts.
However, the associated plots are often simple and do not adequately test domain knowledge or how to find evidence within larger amounts of data.
Our dataset tests domain-specific claim verification abilities over heterogeneous, information-rich figures.

\section{Creating \ourdata}
\label{sec:data_creation}

\begin{figure*}[!t]
    \centering
    \includegraphics[width=\textwidth]{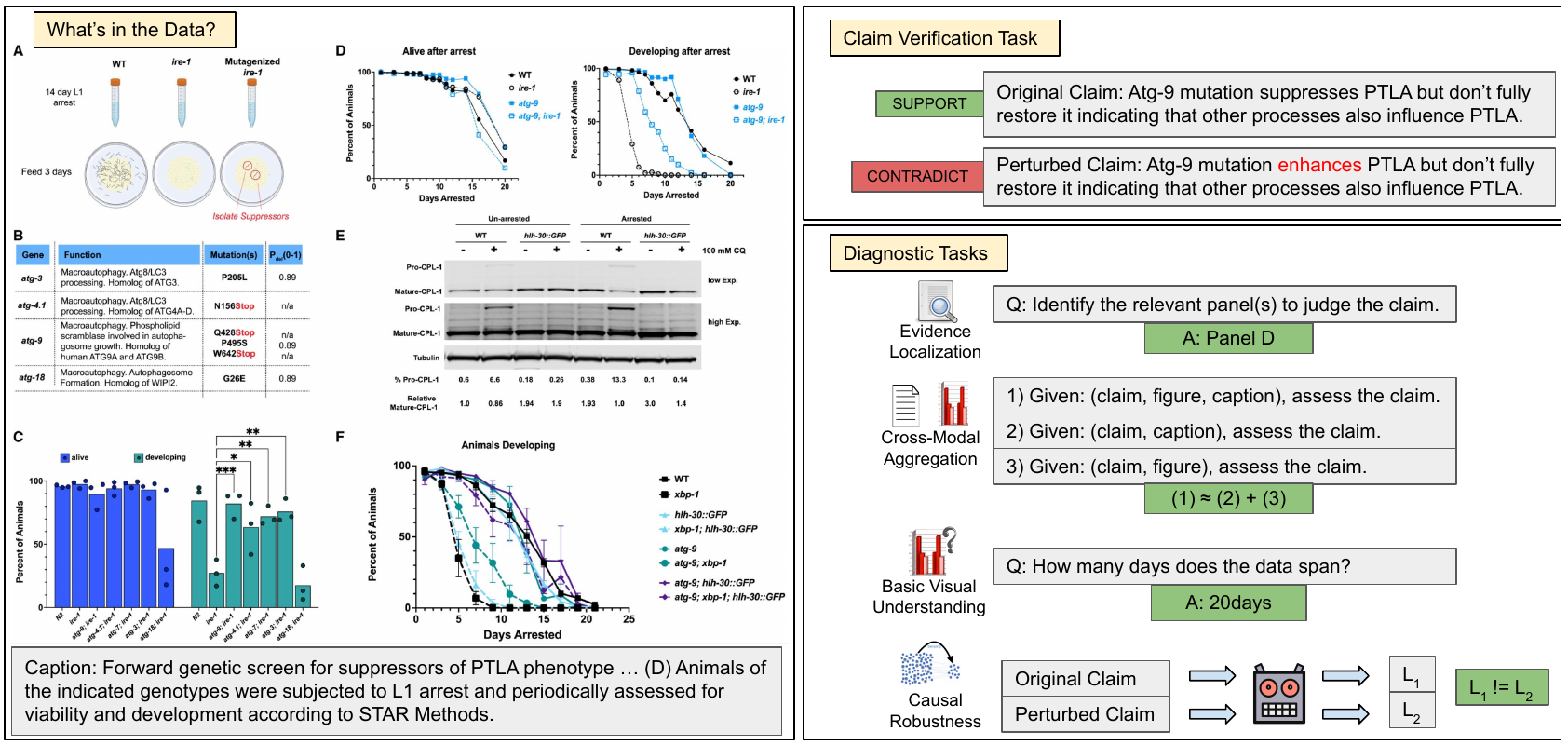}
    \caption{Each data point from \ourdata contains a claim, its associated figure and caption and annotations about its relevant panels. Each claim, both original and perturbed, is also labeled with its relationship to the figure (\support, \neutral, \contradict). It also enables performing diagnostic tests. Models must identify the relevant panel as part of \loc and answer a question about the figure for \bv. To perform \cma, model performance should drop when given either just the figure or the caption. For \rob, model predictions must change across a pair of \texttt{(original, perturbed)} claims. 
    }
    \label{fig:data-example}
\end{figure*}

Verifying whether a claim is supported by scientific evidence requires understanding different parts of the claim, locating and extracting information from multimodal sources, reasoning with it, and finally making a judgment.
In scientific articles, such evidence is often presented graphically, in panels of information-rich figures along with a descriptive caption.
To test whether models can verify claims over scientific figures, we need claims that are supported, as well as ones that are not.
The former can be extracted from papers but manual intervention is required to create the latter. 
To go beyond standard quantitative benchmarking and better understand model failures, we also need tests that are diagnostic in nature.
To this end, we introduce \ourdata, a dataset created from scientific articles across physics, chemistry and biology.
We use open-access, peer-reviewed articles published in Nature Physics, Journal of the American Chemical Society, and Cell, as the source for claims in physics, chemistry and biology respectively.

\subsection{Automatic Extraction}

We extract figures and associated claims from the results section of the articles, where key findings and takeaways are described along with supporting evidence, often expressed within heterogeneous figures containing charts, microscopy images, chemical reaction schemes, or other diagrams. 

Figures in these articles are quite diverse--they vary in size, resolution, placement, and caption style. 
We make use of both the HTML and PDF versions of the articles to obtain a uniform organization and representation of all figures.
We ensure that only high-resolution images (300ppi) are retained and preprocess the captions to remove irrelevant information such as structural prefixes. 

To extract claims associated with the figures, we process the \textit{Results} section text. We use simple regular expressions to identify sentences which either contain explicit references to figures (e.g. ``Fig.'') or some form of inline references (e.g., ``Author et al.''). 
We discard sentences referring to multiple figures or supplementary figures.
Full details of the extraction process are provided in \autoref{appsec:auto_extract}.

\subsection{Systematic Claim Perturbation}
\label{subsec:sys_perturb}

The claims extracted from the articles are grounded in the associated figures---i.e., the figures support\footnote{We rely on the scientific integrity of the published articles and assume that evidence support the asserted claims.} the claims. 
To create an effective test bed, we also need claims that are \emph{not supported} and have \emph{contradictory} evidence in the figures. 
Further, we want to ensure that we test for a variety of reasons that could make a claim unsupported or contradictory with respect to the given evidence. 
To this end, we use a manual claim perturbation process to ensure meaningful perturbations, and ensure their quality through a second annotation process. 
Annotators produce free-form contradicting claims over a range of perturbations.

We manually analyzed the original claims to identify the main capabilities needed for checking a claim against the corresponding figures. 
Based on this, we create four categories of perturbations: 
(i) \emph{Qualitative Inference}---Directional terms are replaced with their opposites (e.g., ``high concentration" to ``low concentration"). 
This tests whether models can check if the asserted qualitative statements are supported via visual relationships between data points in a figure. 
(ii) \emph{Qualitative Relationship Inference}---Comparisons are edited (e.g., ``X is stronger than Y" into ``Y is stronger than X") to create the opposite conclusion from a figure. 
This checks for assessing qualitative relationships between variables via visual inference of similar relationships, 
(iii) \emph{Quantitative reasoning}---Numerical values, primarily associated with experiment details such as statistical significance or experiment size, are modified to test for reasoning about the key quantities of interest.
(iv) \emph{Epistemic Mismatch}---This represents a disconnect between different forms of knowledge.
We add perturbations that introduce inconsistencies between an observation that is visually true (i.e., supported by the figure) and its inference (which requires domain knowledge).
This tests for the ability to carefully connect visually verified information with the textually asserted effect.

Our perturbations ensure that the modified claim is a contradiction of the supported claim. 
This means that the figure which supports the original claim will, by extension, not support the modified claim.
We verify the quality of the resulting perturbations through a second round of manual annotation.
For a subset of data, three annotators were provided a supported claim as well as its perturbation.
They are required to judge whether the perturbation contradicts the supported claim.
We find that all three annotators agree that all perturbed claims are indeed contradictions of the corresponding supported claims (100\% agreement).

\subsection{Diagnostics}
\label{subsec:diagnostics}

\ourdata is also designed to be a diagnostic dataset to support deeper understanding of model capabilities.
We introduce four kinds of diagnostic tests that relate to different aspects of the claim verification problem. 

(i) \bv---For each claim, we identify a data point that is integral to it, and then introduce questions that test for model ability to read or extract it from the figure.
(ii) \loc---The dataset contains automatically extracted annotations about the panels of a figure which should be perused to judge a claim.
Using this, we can test a model's \loc ability in the visual modality.
(iii) \cma---Often, information (such as that about statistical significance) in the caption is also important to correctly assess a claim.
Textual reasoning over the caption must be combined with visual reasoning over the figure.
We test models' multimodal reasoning abilities through \cma.
(iv) \rob---Claims often contain an observation from a figure, as well as an inference that also requires domain knowledge to understand.
Relationships between observations and inferences are systematically perturbed by annotators as part of \S\ref{subsec:sys_perturb}.
We collect annotations about whether it is the observation, the inference, or both, that are perturbed.
Through \rob, we establish how models change their judgments for such perturbations, indicating their understanding of epistemic relationships within claims.
Examples of these diagnostics are provided in \autoref{fig:data-example}.

\subsection{Dataset Statistics} 

Through the process described above, we obtain 505 claims that are supported by a figure (\support), and 505 corresponding perturbed claims in contradiction to a given figure (\contradict).
Further, we pair each claim with an unassociated figure from the same paper to obtain data where there is no connection between them (\neutral).
Therefore, \ourdata contains 1515 data points balanced equally across 3 class labels.
Out of this, 918 data points are from biology, 309 from chemistry and 288 from physics.
Each data point is also annotated with figure panels most relevant to a claim, a question about the figure and information about perturbation types to support our diagnostic tests\footnote{We will release the data in accordance with the papers' CC BY 4.0 license.}.

\begin{table*}[!t]
\centering
\small{
\begin{tabular}{ccrrrrrrrrrrrrr}
\toprule
\multicolumn{2}{l}{} & \multicolumn{3}{c}{\support} & \multicolumn{3}{c}{\neutral} & \multicolumn{3}{c}{\contradict} & \multicolumn{3}{c}{\overall} \\
\cmidrule(lr){3-5} \cmidrule(lr){6-8} \cmidrule(lr){9-11} \cmidrule(lr){12-14}
 & & \multicolumn{1}{c}{P} & \multicolumn{1}{c}{R} & \multicolumn{1}{c}{F} & \multicolumn{1}{c}{P} & \multicolumn{1}{c}{R} & \multicolumn{1}{c}{F} & \multicolumn{1}{c}{P} & \multicolumn{1}{c}{R} & \multicolumn{1}{c}{F} & \multicolumn{1}{c}{P} & \multicolumn{1}{c}{R} & \multicolumn{1}{c}{F} \\
\midrule
\multirow{2}{*}{\gfmini} & \dexpt & 0.41 & 0.88 & 0.56 & 0.64 & 0.48 & 0.55 & 0.75 & 0.10 & 0.17 & 0.60 & 0.48 & 0.43 \\
 & \rdexpt & 0.43 & 0.83 & 0.56 & 0.64 & 0.47 & 0.54 & 0.62 & 0.20 & 0.30 & 0.56 & 0.50 & 0.47 \\
\midrule
\multirow{2}{*}{\gfomni} & \dexpt & 0.43 & 0.93 & 0.59 & 0.86 & 0.46 & 0.60 & 0.75 & 0.23 & 0.35 & 0.68 & 0.54 & 0.51 \\
 & \rdexpt & 0.47 & 0.86 & 0.61 & 0.71 & 0.61 & 0.65 & 0.76 & 0.25 & 0.38 & 0.65 & 0.57 & 0.55 \\
\midrule
\multirow{2}{*}{\sonnet} & \dexpt & 0.52 & 0.87 & 0.65 & 0.83 & 0.64 & 0.72 & 0.79 & 0.43 & 0.56 & 0.71 & 0.65 & 0.64 \\
 & \rdexpt & 0.53 & 0.89 & 0.66 & 0.84 & 0.64 & 0.73 & 0.81 & 0.47 & 0.59 & \textbf{0.73} & 0.66 & 0.66 \\
\midrule
\othree & \rdexpt & \textbf{0.67} & 0.74 & \textbf{0.71} & 0.69 & 0.79 & 0.74 & 0.82 & 0.61 & \textbf{0.70} & \textbf{0.73} & \textbf{0.72} & \textbf{0.72} \\
\midrule
\ofour & \rdexpt & 0.62 & 0.84 & \textbf{0.71} & 0.76 & 0.76 & \textbf{0.76} & 0.88 & 0.57 & 0.69 & 0.75 & \textbf{0.72} & \textbf{0.72} \\
\midrule
\midrule
\multirow{2}{*}{\phimm} & \dexpt & 0.43 & 0.70 & 0.53 & 0.74 & 0.19 & 0.30 & 0.44 & 0.50 & 0.47 & 0.54 & 0.46 & 0.43 \\
 & \rdexpt & 0.36 & 0.81 & 0.51 & 0.81 & 0.10 & 0.17 & 0.58 & 0.26 & 0.36 & 0.58 & 0.41 & 0.34 \\
\midrule
\multirow{2}{*}{\llava} & \dexpt & 0.37 & \textbf{0.92} & 0.53 & 0.68 & 0.33 & 0.44 & \textbf{1.00} & 0.00 & 0.01 & 0.68 & 0.42 & 0.33 \\
 & \rdexpt & 0.38 & 0.80 & 0.52 & 0.54 & 0.42 & 0.47 & 0.61 & 0.09 & 0.15 & 0.51 & 0.43 & 0.38 \\
\midrule
\multirow{2}{*}{\llama} & \dexpt & 0.41 & 0.86 & 0.56 & 0.68 & 0.42 & 0.52 & 0.60 & 0.17 & 0.27 & 0.57 & 0.49 & 0.45 \\
 & \rdexpt & 0.37 & 0.93 & 0.53 & 0.72 & 0.15 & 0.25 & 0.61 & 0.17 & 0.27 & 0.56 & 0.42 & 0.35 \\
\midrule
\multirow{2}{*}{\molmo} & \dexpt & 0.41 & 0.91 & 0.57 & 0.77 & 0.29 & 0.42 & 0.54 & 0.22 & 0.32 & 0.57 & 0.47 & 0.43 \\
 & \rdexpt & 0.39 & 0.75 & 0.51 & 0.56 & 0.28 & 0.37 & 0.43 & 0.23 & 0.30 & 0.46 & 0.42 & 0.39 \\
\midrule
\multirow{2}{*}{\ivl} & \dexpt & 0.62 & 0.79 & 0.70 & \textbf{0.83} & 0.68 & 0.75 & 0.70 & \textbf{0.64} & 0.67 & 0.72 & 0.70 & 0.70 \\
 & \rdexpt & 0.45 & 0.91 & 0.60 & \textbf{0.83} & 0.49 & 0.61 & 0.80 & 0.30 & 0.44 & 0.69 & 0.57 & 0.55 \\
\midrule
\multirow{2}{*}{\qwen} & \dexpt & 0.55 & 0.80 & 0.65 & 0.70 & \textbf{0.80} & 0.75 & 0.85 & 0.34 & 0.48 & 0.70 & 0.65 & 0.63 \\
 & \rdexpt & 0.42 & 0.91 & 0.57 & 0.79 & 0.41 & 0.54 & 0.85 & 0.25 & 0.39 & 0.68 & 0.53 & 0.50 \\
\midrule
\multirow{2}{*}{\ds} & \dexpt & 0.55 & 0.43 & 0.48 & 0.50 & 0.67 & 0.58 & 0.44 & 0.40 & 0.42 & 0.50 & 0.67 & 0.58 \\
 & \rdexpt & 0.41 & 0.65 & 0.51 & 0.51 & 0.51 & 0.51 & 0.51 & 0.21 & 0.30 & 0.48 & 0.46 & 0.44 \\
\bottomrule
\end{tabular}
}
\caption{
Model performance on the claim verification task of \ourdata when prompted to simply generate the decision (\dexpt), and when asked to reason and then generating the decision (\rdexpt). \ivl achieves best performance when prompted to just give the answer, while \othree and \ofour are the best overall, using their inbuilt reasoning capabilities. Closed-source models are slightly better with reasoning whereas open-source models do worse in most cases, represent a significant gap in their reasoning capabilities.}
\label{tab:main-results}
\end{table*}

\section{Experimental Setup}
\label{sec:expts}

We benchmark the performance of several state-of-the-art vision-language models (\vlms) on evaluation tasks supported by \ourdata.

\subsection{Evaluation Tasks}

\ourdata is designed as a \ver task.
Each data point contains a claim, an associated (multi-panel) figure (and caption) and a label (\support, \neutral, \contradict).
Given the figure (and caption) and a claim, models must generate a prediction about whether the claim is supported.
We evaluate models on this task using standard metrics of precision, recall and F1 score.

\ourdata also supports four \emph{diagnostic} tasks designed to assess a diverse set of capabilities required to effectively verify claims.
Performance on these diagnostics highlight limitations of contemporary models, thereby opening up avenues for future research.
(1) \loc tests whether models can localize to the correct panel(s) in the figure.
Given the figure (and caption) and a claim, models must generate the relevant panel names as well as generate a prediction (\ver).
We use precision, recall and F1 to measure how well models identify the correct panels.
(2) \bv aims to test whether models can read scientific figures by how models answer a question about the figure.
Each claim in \ourdata is accompanied by a basic question and its (one-word) answer about the associated figure.
We use Exact Match to judge whether a model answer is correct.
(3) \cma are experiments designed to analyze how models use the figure and its caption to come up with their judgment.
Models need to aggregate information from the figure (visual information) as well as caption (textual information) for claim verification.
First, models are given a claim, the associated figure, its caption and required to perform \ver.
Then, for the same task, they are prompted to reason over just the figure and just the caption, testing its visual and textual abilities respectively.
(4) \rob tests whether models consistently (and correctly) change their prediction across epistemic perturbations of the same claim; they should predict support for the original and contradict for the perturbed claim. 
Claims often encode epistemic information---observations from the figure and related inferences made with domain knowledge.
As part of \S\ref{subsec:sys_perturb}, annotators also mark whether they perturb the observation, the inference or both.
We perform a sensitivity test for the same across \texttt{(original, perturbed)} claim pairs.

\subsection{Models}
\label{subsec:models}

We conduct the aforementioned evaluation on a set of $12$ different vision-language models (\vlms): gpt-4o-mini-2024-07-18 (\gfmini), gpt-4o-2024-11-20 (\gfomni), claude-3-5-sonnet-20241022 (\sonnet), o3-2025-04-16 (\othree), o4-mini-2025-04-16 (\ofour), Phi-4 Multimodal Instruct (\phimm), llava-v1.6-mistral-7b-hf (\llava), Llama-3.2-11B Vision Instruct (\llama), Molmo-7B-D (\molmo), InternVL3-38B (\ivl), Qwen2.5-VL-32B (\qwen) and deepseek-VL2-small (\ds). 
This set represents both open and closed-sourced models of differing capabilities for a comprehensive evaluation of \ourdata.
We evaluate models primarily in two zero-shot settings: (i) generating only a judgment (\dexpt), and (ii) reasoning about the claim before judging it (\rdexpt).
More details are in \autoref{appsec:benchmark_models}, \ref{appsec:prompts} and \ref{appsec:model_setup}.

\section{Results}

\autoref{tab:main-results} presents the performance of all the models in different settings for the multimodal scientific claim verification task. 
We present per-class (\support, \neutral and \contradict) precision, recall and F1 score as well as macro average metrics on the class balanced \ourdata\footnote{Results for each domain are presented in \autoref{tab:phy-results}, \ref{tab:chem-results} and \ref{tab:bio-results}}. 
We make two main observations.

\paragraph{Most \vlms perform poorly on \ourdatasmall.}
We observe that most models perform poorly on the task (\dexpt rows in \autoref{tab:main-results}), with overall F1 scores only ranging from $\sim$0.3-0.5. 
Only two models (out of ten) stand out: \sonnet (0.66 F1) and \ivl (0.70 F1). 
Models attain high recall and low precision on \support. In contrast for the \neutral and \contradict claims, models have low recall and high precision. This implies two findings: First, models have a strong bias towards recognizing most of the claims as supported. Second, the models can reliably identify some of the \neutral and \contradict claims. Our manual analysis shows that models only identify the most obviously wrong claims as the \neutral and \contradict claims. This can be done reliably but they struggle on ones that are more difficult, which require careful reasoning. These suggest challenges for claim verification methods.

\paragraph{Reasoning before judging helps models slightly.}
The \rdexpt rows in \autoref{tab:main-results} show results where models, given the figure and caption, first perform step-by-step reasoning on the claim and then generate their decision on the category of the claim. 
Results show that reasoning leads to improvements ($\sim$0.02-0.04) for closed-source models and \llava, but the gains are rather small.
\othree and \ofour, models trained to analyze and do reasoning over images, achieve the highest performance (0.72 F1).

There is a notable drop in performance for open-source models ($\sim$0.04-0.16) indicating a weakness in CoT abilities of open-source models for claim verification.
We hypothesize that this is due to the limitations of instruction tuning in vision-language modeling where models are mainly finetuned to describe or analyze images, not reasoning chains. 

\autoref{tab:domain-results} presents model performance for different domains in \ourdatasmall.
On average,  models are worst at verifying \phy claims and best at judging \chem claims.
However, the highest performance is achieved on \bio claims.

\section{Diagnostics Results}

We use our diagnostic tests to better understand the failure modes of \gfmini, \gfomni, \sonnet and \ivl.
Going forward, we run these diagnostic tests (\S\ref{subsec:diagnostics}) on the \bio subset of \ourdata and discuss them.

\begin{table}[!ht]
\centering
\begin{tabular}{ccrrr}
\toprule
 &  & \multicolumn{1}{c}{P} & \multicolumn{1}{c}{R} & \multicolumn{1}{c}{F} \\
\midrule
\multirow{2}{*}{\gfmini} & \rdexpt & 0.57 & 0.50 & 0.46 \\
 & \irdexpt & 0.59 & 0.45 & 0.40 \\
\midrule
\multirow{2}{*}{\gfomni} & \rdexpt & 0.69 & 0.59 & 0.56 \\
 & \irdexpt & 0.73 & 0.51 & 0.47 \\
\midrule
\multirow{2}{*}{\sonnet} & \rdexpt & 0.78 & \textbf{0.70} & \textbf{0.70} \\
 & \irdexpt & \textbf{0.79} & 0.69 & \textbf{0.70} \\
\midrule
\multirow{2}{*}{\ivl} & \rdexpt & 0.75 & 0.59 & 0.58 \\
 & \irdexpt & 0.75 & 0.59 & 0.58 \\
\bottomrule
\end{tabular}
\caption{Model performance on claim verification worsens when also prompted to localize to the relevant panels (\irdexpt) as compared to reasoning over the entire figure and assessing a claim (\rdexpt).}
\label{tab:localization_performance}
\end{table}

\begin{table}[!ht]
\centering
\begin{tabular}{crrr}
\toprule
 & \multicolumn{1}{c}{P} & \multicolumn{1}{c}{R} & \multicolumn{1}{c}{F} \\
\midrule
\gfmini & 0.37 & 0.77 & 0.50 \\
\midrule
\gfomni & 0.53 & 0.70 & 0.61 \\
\midrule
\sonnet & \textbf{0.62} & \textbf{0.80} & \textbf{0.70} \\
\midrule
\ivl & 0.46 & 0.68 & 0.55 \\
\bottomrule
\end{tabular}
\caption{\loc---We use precision, recall and F1 score to characterize how well models can localize to relevant panels. Low precision indicates that they are bad at identifying only the correct panels. 
}
\label{tab:panel-metrics}
\end{table}

\paragraph{\vlms localize poorly to relevant information.}
Finding the most relevant panels of figures is important to assess claims from information-rich figures. \autoref{tab:localization_performance} shows how well models perform when prompted to first identify the associated panels, reason over them and make a decision (\irdexpt), thereby testing \loc.
Model performance deteriorates when localizing before reasoning (\irdexpt) as compared to reasoning over the entire figure (\rdexpt).
We also explicitly test their ability to locate relevant panels.
\autoref{tab:panel-metrics} presents precision, recall and F1 to measure how well models can localize to the correct visual evidence.
Their low precision and high recall indicates they do identify the relevant panel(s), but also deem a lot of irrelevant panels to be important.
Clearly, evidence localization is difficult for models.

\begin{figure}[!ht]
    \centering
    \includegraphics[width=\columnwidth]{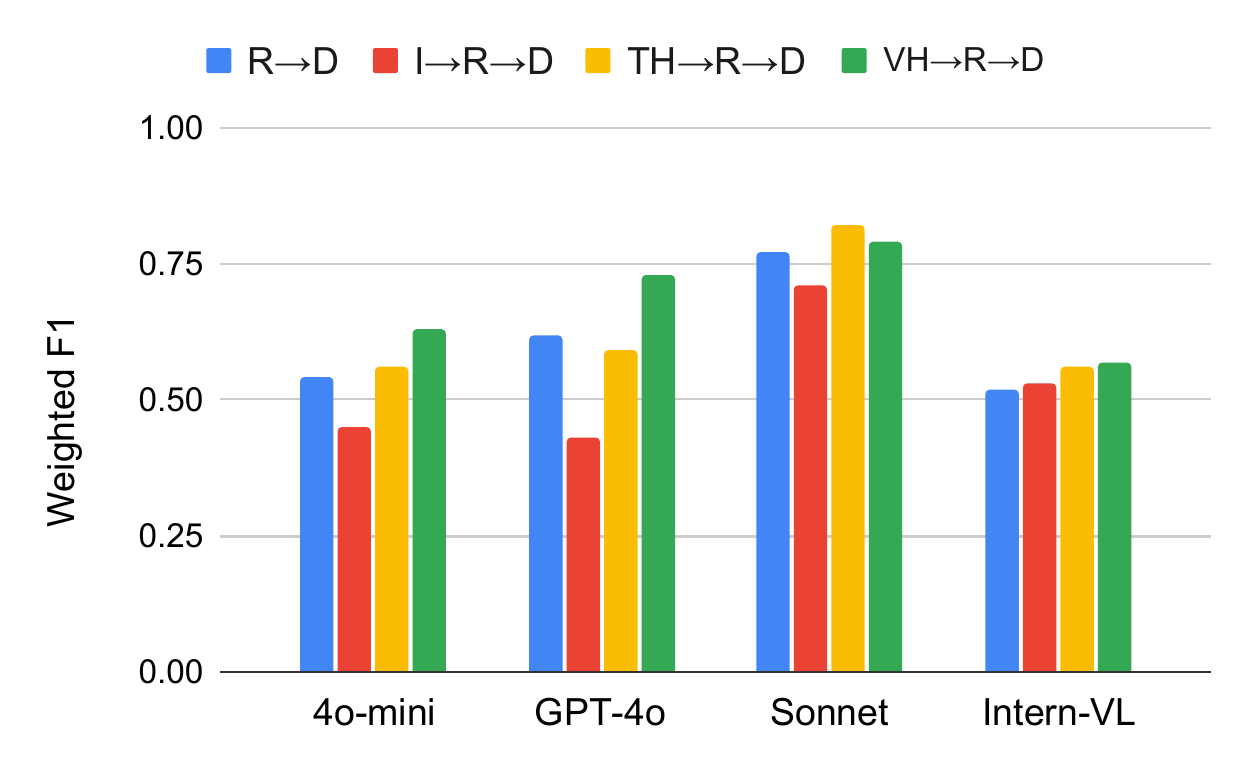}
    \caption{
    Model performance on \ver when prompted for (or provided) localization.
    Providing models with hints about the relevant panels of figures improves their claim verification. Textual hints (TH) guide models to focus on the correct part of the full figure, showing higher performance than \rdexpt and \irdexpt. Models using visual hints (VH; relevant panel as visual input instead of full figure) perform even better. This indicates that localizing to the relevant knowledge has the potential to improve models.
    }
    \label{fig:localization-upside}
\end{figure}

\paragraph{Better localization can improve performance.}
We perform a series of experiments to establish how well models can perform if they have correct localization information.
First, we provide models gold information about which panels are associated with the claim as a \emph{textual hint} (TH$\rightarrow$\rdexpt).
Next, for each claim, instead of the full figure, we only provide the relevant panel to the model as a \emph{visual hint} (VH$\rightarrow$\rdexpt), instead of the full figure.
These experiments are performed over a randomly sampled subset (n=101) of class-balanced data points.

\autoref{fig:localization-upside} compares the performance of models with and without these hints.
As stated earlier, models are better at reasoning over the full figure (\rdexpt) rather than over panels it has identified as relevant (\irdexpt).
However, when given the relevant panels as a textual hint (TH$\rightarrow$\rdexpt), they fare much better.
They improve even further when only given the relevant panel(s) of the figure (VH$\rightarrow$\rdexpt) as input, thus removing panel localization errors.
The poor localization performance coupled with the gains seen with localization hints suggest that improving the localization abilities of models is valuable. 
But even with perfect localization (i.e., through hints here), there is significant room for improvement, indicating challenges in other aspects of multimodal reasoning.

\begin{figure}[!ht]
    \centering
    \includegraphics[width=\columnwidth]{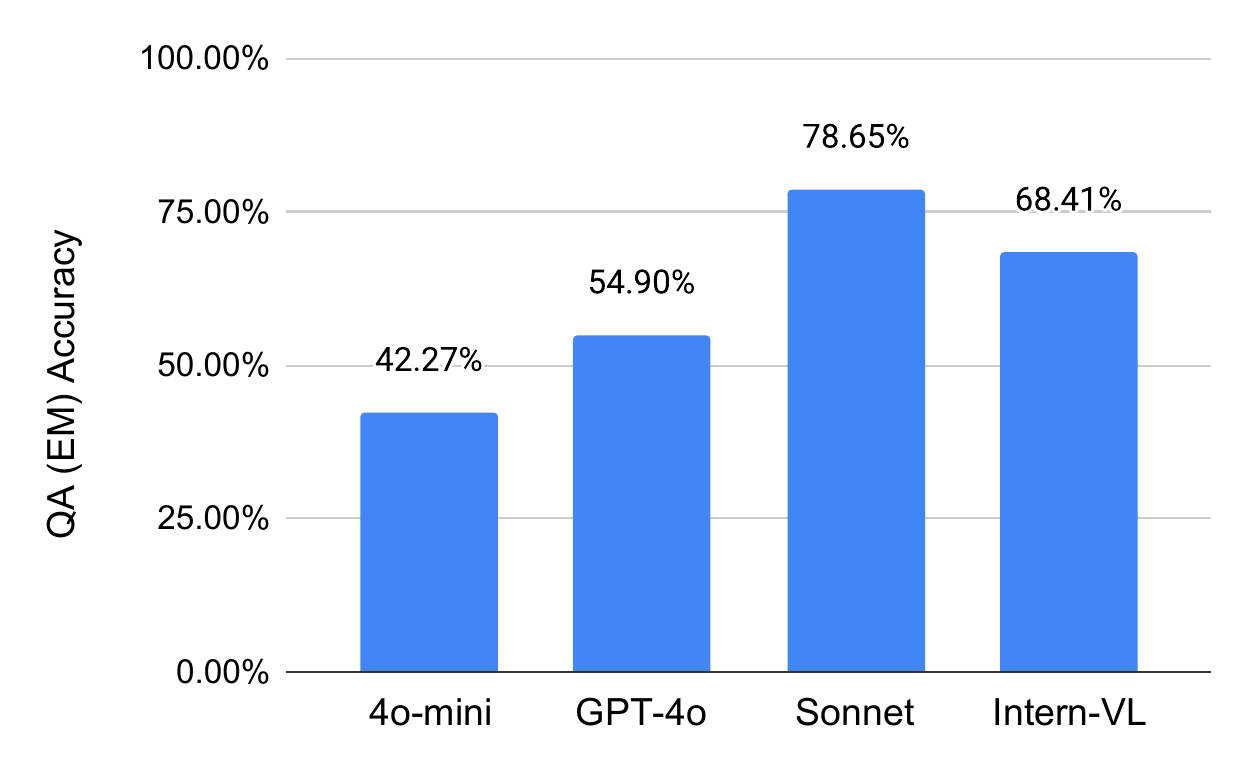}
    \caption{
    Model performance on \bv.
    Models fail to answer basic questions about components of figures associated with claims.}
    \label{fig:basic-vqa}
\end{figure}

\paragraph{Models need to improve on basic visual reading.}
Each claim in \ourdata is also accompanied by a question about the figure that is relevant for the verification process.
These questions test basic visual reading abilities (e.g., ``How many days does the data span?" in \autoref{fig:data-example}) and do not require complex reasoning.
\autoref{fig:basic-vqa} shows that most models perform poorly on such questions.
\sonnet performs the best, correctly answering $\sim$78\% of the questions.
The moderate performance indicates a gap in models' visual comprehension capabilities when it comes to scientific figures.

\begin{table}[!ht]
\centering
\begin{tabular}{ccrrr}
\toprule
 &  & \multicolumn{1}{c}{F+C} & \multicolumn{1}{c}{C} & \multicolumn{1}{c}{F} \\
\midrule
\multirow{2}{*}{\gfmini} & D & 0.42 & 0.46 & 0.38 \\
 & R$\rightarrow$D & 0.46 & 0.50 & 0.45 \\
\midrule
\multirow{2}{*}{\gfomni} & D & 0.52 & 0.50 & 0.45 \\
 & R$\rightarrow$D & 0.56 & 0.44 & 0.51 \\
\midrule
\multirow{2}{*}{\sonnet} & D & 0.68 & 0.58 & 0.60 \\
 & R$\rightarrow$D & 0.70 & 0.51 & 0.64 \\
\midrule
\multirow{2}{*}{\ivl} & D & 0.74 & 0.64 & 0.68  \\
 & R$\rightarrow$D & 0.58 & 0.47 & 0.47 \\
\bottomrule
\end{tabular}
\caption{Models achieve a large chunk of their performance using information from just one modality even though information from both modalities is needed to judge claims. F+C indicates when both the figure and the caption is provided, C indicates when only the caption (textual) is provided, and F indicates when only the figure (visual) is provided to the model.}
\label{tab:cross-modal}
\end{table}

\paragraph{Models struggle with cross-modal reasoning.}
\autoref{tab:cross-modal} compares model performance when provided both the figure and its caption, just the caption and just the figure.
Models must reason over information in both modalities in order to best assess a claim since information is found in both the figure (visual) as well as its caption (textual).
However, we note that models' performance doesn't improve substantially over its performance when using just one modality.
This indicates that they might not be effectively combining the complementary information present in both modalities.

\begin{table}[!ht]
\centering
\begin{tabular}{ccrrrr}
\toprule
 &  & \multicolumn{1}{c}{Obs} & \multicolumn{1}{c}{Inf} & \multicolumn{1}{c}{Both} & \multicolumn{1}{c}{None} \\
\midrule
\multirow{2}{*}{4o-mini} & D & 13\% & 13\% & 0\% & 12\% \\
 & R$\rightarrow$D & 20\% & 28\% & 0\% & 22\% \\
\midrule
\multirow{2}{*}{4o} & D & 13\% & 30\% & 0\% & 23\% \\
 & R$\rightarrow$D & 7\% & 26\% & 0\% & 27\% \\
\midrule
\multirow{2}{*}{Sonnet} & D & 47\% & 50\% & 0\% & 45\% \\
 & R$\rightarrow$D & 47\% & 54\% & 0\% & 52\% \\
\midrule
\multirow{2}{*}{InternVL} & D & 67\% & 72\% & 50\% & 65\% \\
 & R$\rightarrow$D & 60\% & 39\% & 0\% & 35\% \\
\bottomrule
\end{tabular}
\caption{Model sensitivity---changing their prediction about a claim for different types of perturbation.}
\label{tab:causal-robustness}
\end{table}

\paragraph{Models can't handle epistemic mismatches.}
Claims often encode epistemic relationships which can be systematically perturbed to test the sensitivity of contemporary models \cite{verma-etal-2023-evaluating}.
We calculate sensitivity as the percentage of times models change predictions across the supported and refuted version of the same claim.
\autoref{tab:causal-robustness} shows the sensitivity of models by perturbation type.
Models are not sensitive enough to understand nuances in epistemic relationships, being the worst when both the observation and inference is modified.
Analyzing differences in models’ confidences for predictions may provide more insight \cite{marce-poliak-2022-gender}.

\section{Conclusion}

Assessing whether claims are supported requires understanding the methods and data presented in associated figures.
One must find the correct piece of information in the figure and then combine it with the caption.
This paper introduces \ourdatasmall, a new diagnostic dataset to evaluate the claim verification capabilities of \vlms.
We find that most \vlms are poor at this task out-of-the-box, and chain-of-thought only helps slightly.
Particularly, they are significantly worse at understanding that given evidence contradicts (or is not related to) the claim.
\loc shows that models are bad at identifying the right panel of data, a critical flaw in their claim verification capabilities.
\cma indicates that models do not effectively use both visual and textual information for their judgments.
Diagnostics also reveal that they do not understand some obvious characteristics of the associated figures.
Our results establish the current abilities of \vlms for claim verification over heterogeneous, information-rich scientific figures, and our diagnostics highlight specific avenues of research to improve them.

\section*{Limitations}

We benchmark a reasonably diverse set of \vlms.
However, we acknowledge that we can try more models across a spectrum of architectures, training paradigms and sizes.
Due to the current fast-paced landscape of \vlm development, we will continue to evaluate more \vlms on \ourdata.

Due to the difficulty of creating data that \vlms have not already seen (published after their cutoff dates), we are unable to train models for this task.
We perform few-shot experiments with closed-source models (\gfmini, \gfomni and \sonnet) but we leave further exploration of different methods of example selection to future work.

We formulate the task of multimodal scientific claim verification.
But our dataset is limited to using captions as the textual part of the input to models.
While these captions are descriptive, models might benefit from using extra context, such as that extracted from the \textit{Methods} sections of papers.

In this work, we perform some qualitative evaluation of weaknesses in the reasoning produced by \vlms.
However, we are unable to do so at scale.
Such evaluation requires experts with incredibly specific domain expertise.
Even a graduate student (PhD level) or faculty cannot verify reasoning for all domains covered in Nature Physics, Journal of the American Chemical Society and Cell.
For instance, an expert in ecology cannot easily judge the reasoning about claims in cellular biology.
We will explore how to conduct better evaluations as part of future work.

Our work only investigates English-language documents and this limits the generalizability of our findings to other languages, although most scientific articles are disseminated in English.

Due to high cost of the recently released \othree and \ofour models, we are unable to analyze it across the full spectrum of our diagnostics.
For consistency, we analyze \sonnet and \ivl since they have similar performance on \ourdata.

\section*{Ethical Considerations and Risks}

Prior work has shown that \vlms exhibit various types of bias.
While they do not generate free-form language for our binary prediction task, it is possible, though highly unlikely, that biases explicitly come up in the explanations.
Deploying such unreliable models into critical infrastructure and relying on them for decisions can cause harm to users.

\section*{Acknowledgments}

This material is based on research that is in part supported by the DARPA for the SciFy program under agreement number HR00112520301. The U.S. Government is authorized to reproduce and distribute reprints for Governmental purposes notwithstanding any copyright notation thereon. The views and conclusions contained herein are those of the authors and should not be interpreted as necessarily representing the official policies or endorsements, either express or implied, of DARPA or the U.S. Government.
The authors would also like to thank Aakanksha Rajiv Kapoor for her help with understanding the structure and content of papers published in the Cell journal and for help with the qualitative error analysis of model outputs.

\bibliography{custom, anthology}

\clearpage

\appendix

\section{Benchmark Models}
\label{appsec:benchmark_models}

We provide details of each model we evaluate on \ourdata.

\paragraph{\texttt{gpt-4o-2024-11-20}} accepts as input any combination of text, audio, image, and video and generates any combination of text, audio, and image outputs.
It is especially better at vision and audio understanding compared to existing models.

\paragraph{\texttt{gpt-4o-mini-2024-07-18}} has a context window of 128K tokens, supports up to 16K output tokens per request.
It surpasses other small models released to that date on academic benchmarks across both textual intelligence and multimodal reasoning, and supports the same range of languages as \gfomni.

\paragraph{\texttt{claude-3-5-sonnet-20241022}} sets new industry benchmarks for graduate-level reasoning (GPQA), undergraduate-level knowledge (MMLU), and coding proficiency (HumanEval). 
It shows marked improvement in grasping nuance, humor, and complex instructions, and is exceptional at writing high-quality content with a natural, relatable tone. 

\paragraph{\texttt{o3-2025-04-16}} excels at solving complex math, coding, and scientific challenges while demonstrating strong visual perception and analysis. It uses tools in its chains of thought to augment its capabilities; for example, cropping or transforming images, searching the web, or using Python to analyze data during the thought process.

\paragraph{\texttt{o4-mini-2025-04-16}} is a smaller model optimized for fast, cost-efficient reasoning—it achieves remarkable performance for its size and cost, particularly in math, coding, and visual tasks. It is the best-performing benchmarked model on AIME 2024 and 2025. It performs especially strongly at visual tasks like analyzing images, charts, and graphics.

\paragraph{\texttt{Phi-4-multimodal-instruct}} is a 5.6 billion parameter multimodal modal that combines image, textual and audio modalaties into a single small language model via LoRA adapters and modality-specific routers that make multiple inference modes possible without interference. The model has been extensively instruction tuned on a combination of synthetic and web data.

\paragraph{\texttt{llava-v1.6-mistral-7b-hf}} is a 7.6 billion parameter vision language model that is part of the Llava-Next regimen and built on top of the Llava architecture. It has a pretrained vision encoder and Mistral-7B as the language modeling backbone. It has been instruction tuned on over a million data points coming from a combination of high-quality user instruct data and multimodal document/chart.

\paragraph{\texttt{Llama-3.2-11B-Vision-Instruct}} is the 11B version of the Llama 3.2-Vision set of multimodal LLMs which have been instruction tuned for image reasoning. It is built on top of the pretrained Llama 3.1 text only LLM by combining a seperately trained vision adapter module. Using a combination of supervised fine-tuning and reinforcement learning from human feedback, the model has been optimized to do a variety of vision tasks like image recognition, reasoning, captioning, and question answering on images.

\paragraph{\texttt{Molmo-7B-D-0924}} is a 7 billion parameter open-source vision-language model. It is developed upon the Qwen2-7B language model with OpenAI CLIP as the vision adapter. The model has been trained on PiXMo, a dataset containing 1 million high quality curated (image,text) tuples.

\paragraph{\texttt{InternVL3-38B}} is a 38 billion parameter open-source vision language model. It has been built based upon the following components: variable visual position encoding which handles longer multimodal context; native multimodal pre-training that combines language pre-training and multimodal post-training in a single pipeline; mixed preference optimization to align the model response distribution with the ground-truth distribution; and test-time scaling using VisualPRM-8B as a critic model for Best-of-N evaluation.

\paragraph{\texttt{Qwen2.5-VL-32B-Instruct}} is a 32 billion parameter vision language model. It is created on top of the Qwen-2.5 7 billion language model by following the ViT architecture. It has been extensively instruction tuned on (image,text) tuples to so that the model understands all things visual, is agentic, can comprehend long videos and events, can do visual localization, and generate structured outputs.

\paragraph{\texttt{deepseek-vl2-small}} is a 16 billion parameters mixture-of-experts vision language model. It has shown been to demonstrate enhanced performance across multiple tasks like visual question answering, optical character recognition, document/table/chart understanding, and visual grounding. It improves upon its predecessor, DeepSeek-VL, by using an improved high-resoultion vision encoder for better visual comprehension and an optimized language model backbone for training and test time efficiency. It is trained on a data that boosts performance and gives new capabilities to the model such as precise visual grounding.

\section{Human Annotation Details}

\begin{figure*}[!t]
    \centering
    \includegraphics[width=\textwidth]{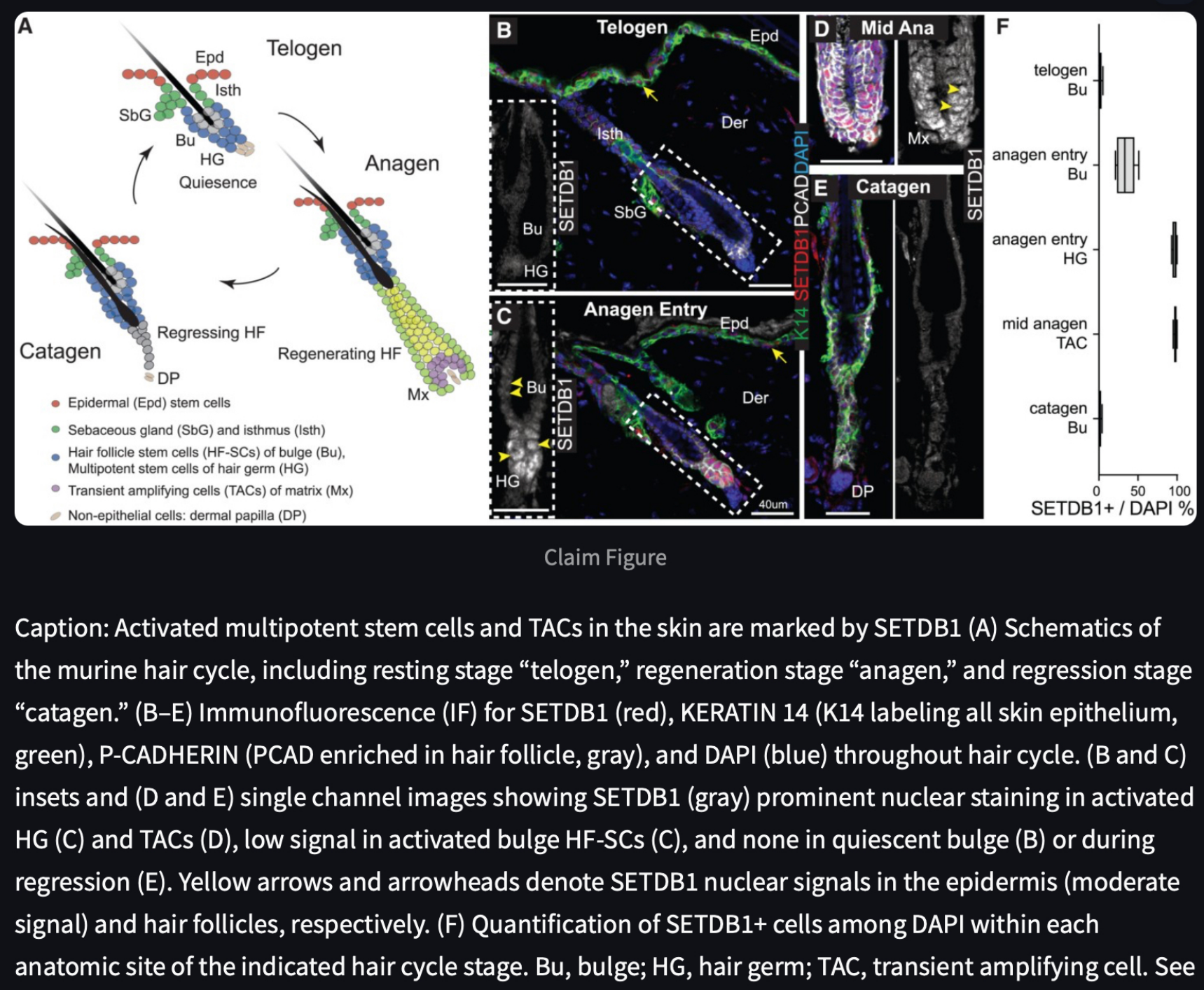}
    \caption{Instructions and UI of the application used to collect perturbations of claims from manual annotators. 
    }
    \label{fig:perturbation-1}
\end{figure*}

\begin{figure*}[!t]
    \centering
    \includegraphics[width=\textwidth]{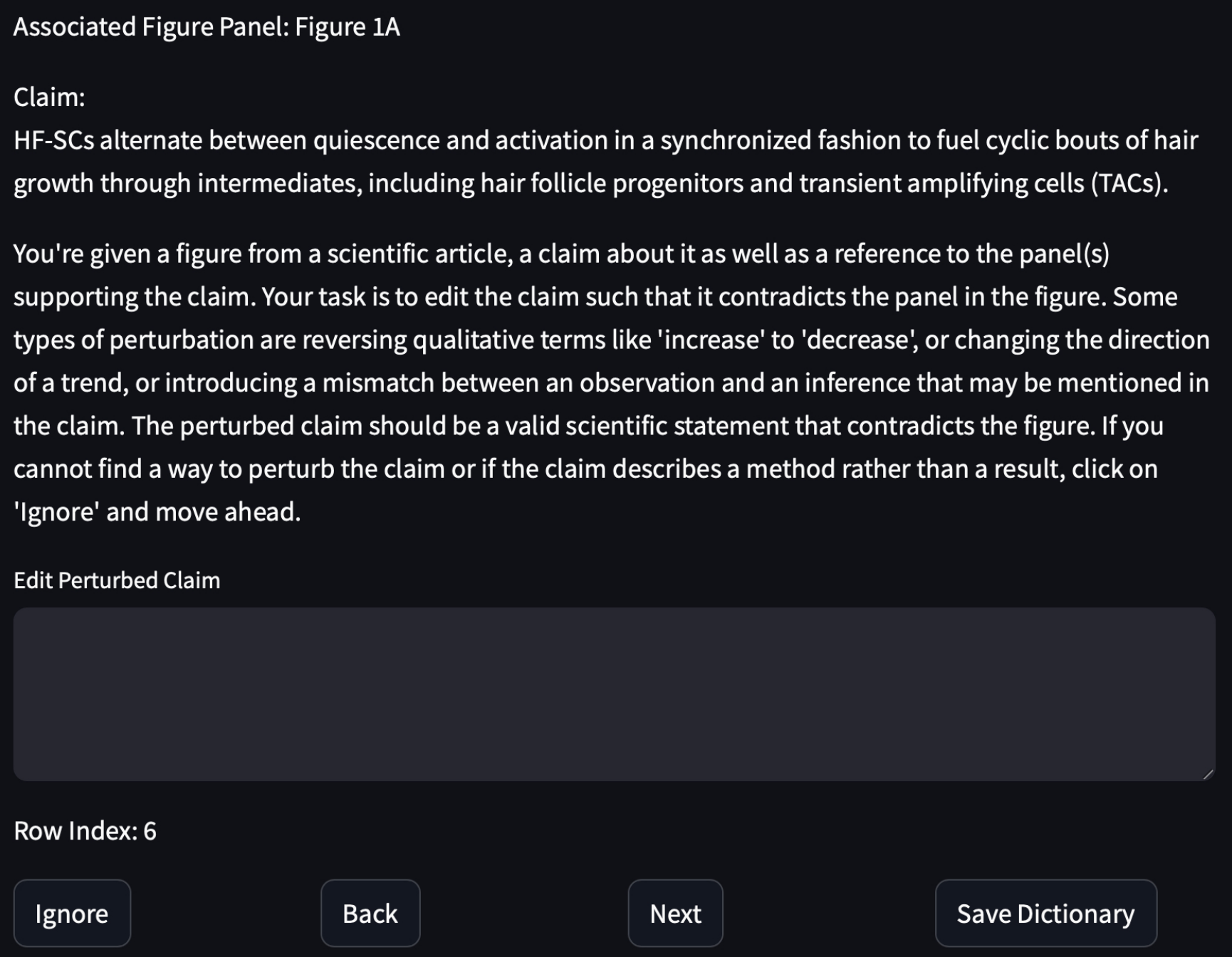}
    \caption{Instructions and UI of the application used to collect perturbations of claims from manual annotators (contd.). 
    }
    \label{fig:perturbation-2}
\end{figure*}

\begin{figure*}[!t]
    \centering
    \includegraphics[width=\textwidth]{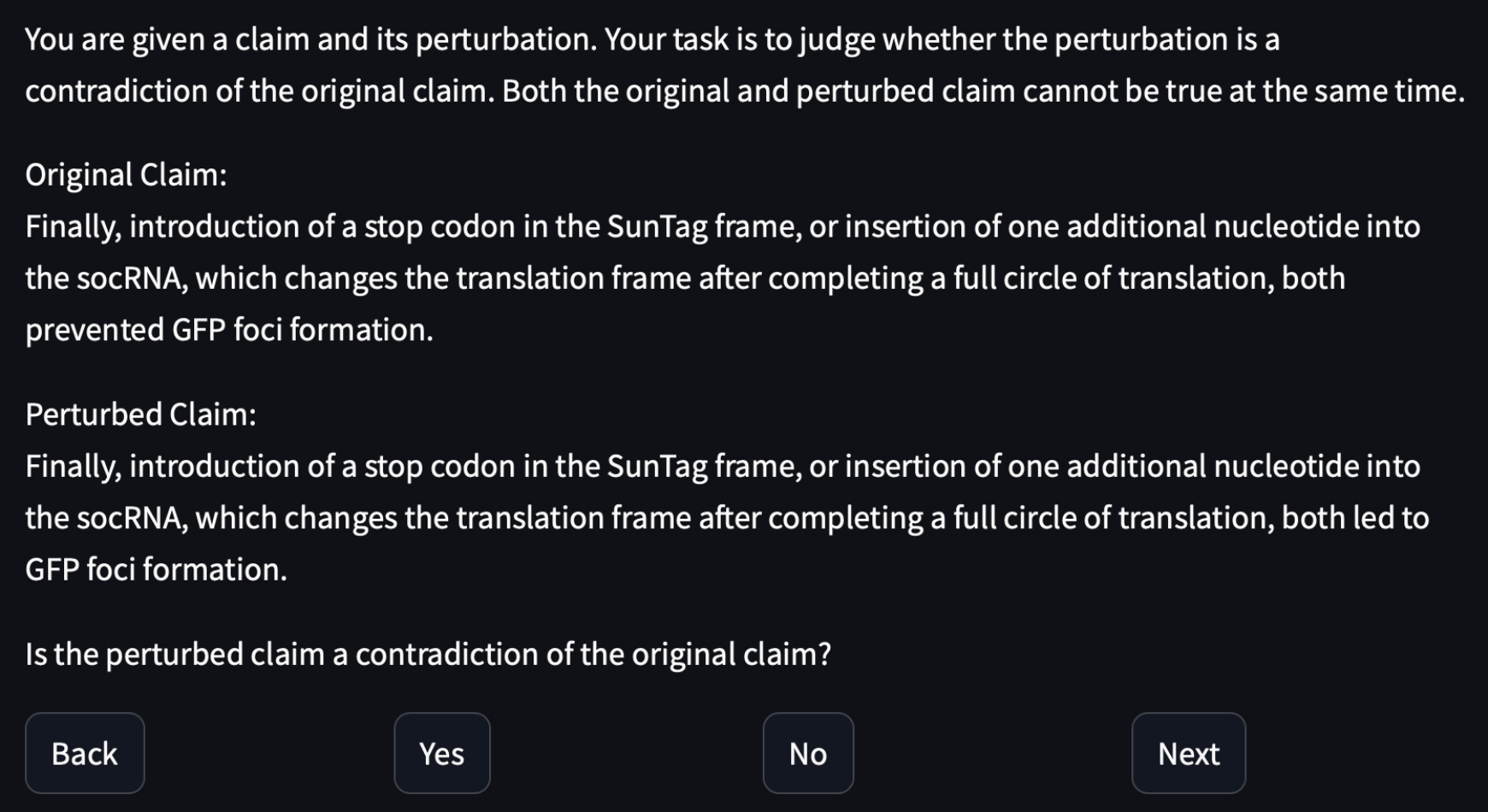}
    \caption{Instructions and UI designed to collect a second round of manual annotation to verify that the perturbed claims are contradictions of the supported claims through \autoref{fig:perturbation-1} and \autoref{fig:perturbation-2}. We ask three annotators to do this task and find that they full agree for all the perturbations. 
    }
    \label{fig:second-manual-annotation}
\end{figure*}

The design and instructions for the different applications through which annotations are collected can be found in \autoref{fig:perturbation-1}, \ref{fig:perturbation-2} and \ref{fig:second-manual-annotation}.
Our annotators are graduate students who are experts at reading figures but have limited domain knowledge.
To alleviate the skill issue, we ask them to avoid perturbations that require more domain knowledge than they possess. 
They were not paid for annotations, and were informed of how the annotations would be used.

\section{Automatic Extraction Details}
\label{appsec:auto_extract}

To construct a high-quality multimodal benchmark for scientific claim verification, we developed an automated pipeline for extracting textual claims and their associated visual elements from research articles. 
Our approach operates over full-text HTML and PDF documents sourced from Nature Physics (\url{https://www.nature.com/nphys/}), the Journal of the American Chemical Society (\url{https://pubs.acs.org/journal/jacsat}) and Cell (\url{https://www.cell.com}), leading impact factor venues in their respective fields.
It creates reliable mapping between complex scientific assertions and their supporting visual evidence with minimal manual supervision.
During dataset collection, no personal identifying info (PII) was collected. None of the collected data contained any offensive content.

\begin{figure}[!htb]
    \centering
    \includegraphics[width=\columnwidth]{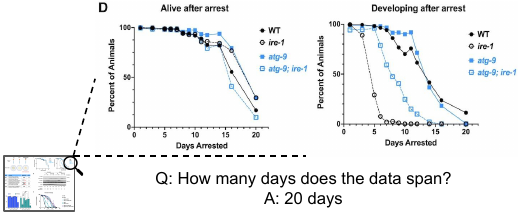}
    \caption{Each claim is accompanied by a diagnostic question that tests whether models can read the relevant panel of the claim's associated figure.}
    \label{fig:diagnostic-question-example}
\end{figure}

\subsection{Automatic Figure Extraction}
\label{appsubsec:fig_extract}

We extract figures and their corresponding captions from the structured HTML versions of articles sourced from these journals. 
Each article contains embedded figure blocks that follow consistent filename conventions and DOM structures, allowing for reliable identification and extraction. 
Figures are mapped to canonical identifiers (e.g., \texttt{figure\_1}, \texttt{figure\_2}, etc.) to ensure consistency across the dataset.

Captions are extracted from the \texttt{<figcaption>} elements associated with each figure and typically consist of a short title followed by descriptive text.
We concatenate these segments, remove structural prefixes and apply light normalization to clean residual markup or formatting noise. 
Only high-resolution main-text figures are retained, while supplementary or non-standard assets are excluded.
This approach yields a clean, structured mapping between each visual element and its corresponding caption, enabling precise alignment with textual claims during the dataset construction process.

\subsection{Automatic Claim Extraction}
\label{appsubsec:claim_extract}

Scientific claims are typically concentrated in the \textit{Results} section, where authors present novel findings grounded in empirical data, often accompanied by figures such as charts, microscopy images, or diagrams. 
In contrast, other sections such as \textit{Introduction} or \textit{Discussion} tend to be more speculative, summarizing prior work or offering high-level interpretations. To ensure that extracted statements are factual, visually grounded, and suitable for verification, we restrict claim extraction to the \textit{Results} section.

We process article PDFs using a layout-aware parser to identify the \textit{Results} section and extract its contents. 
Section headers such as ``Results'' and ``Discussion'' are detected using regex patterns robust to formatting variations and numbering conventions. 
The extracted text is segmented into candidate sentences using a customized version of the NLTK Punkt tokenizer, adapted for scientific prose by accounting for common abbreviations (e.g., ``Fig.'', ``et al.'') and inline structures such as references and equations.

Candidate sentences are filtered using a series of quality criteria to ensure that only concise, visually grounded claims are retained. 
Specifically, each sentence must (i) contain an explicit reference to a main-text figure (e.g., ``(Figure 2A)''), (ii) be between 40 and 800 characters in length, (iii) include at least 8 words, and (iv) not match known patterns associated with citations, table fragments, or supplementary material. To maintain clarity and reduce ambiguity during alignment, we retain only single-sentence claims that refer to a single primary figure (i.e., claims with multiple distinct figure references are excluded). 
Additionally, we restrict figure selection to images smaller than 5MB to ensure compatibility with downstream modeling.
This process yields a clean set of scientific claims, each grounded in a single visual source and suitable for fine-grained multimodal verification and localization tasks.

\section{Model Setup}
\label{appsec:model_setup}
Python was the main scripting language for data collection and experimentation.
For experiments using closed source models, we used OpenAI\footnote{\url{https://openai.com/api/pricing/}} and Anthropic\footnote{\url{https://://www.anthropic.com/pricing}} APIs. The total cost for OpenAI was $\sim$ 400 USD and  $\sim$ 160 USD for Claude. 
The open-source experiments were conducted on $4$ A6000 GPUs, each having $48$ GB. The total GPU hours for all the experiments was $\sim$40. 
The models were downloaded from Huggingface and hosted for inference using Huggingface transformers module and vLLM. 
We use GitHub Co-Pilot to help with writing code but verify it manually before running any experiments.

\section{Additional Results}

\subsection{Model Performance by Domain}
\label{appsec:domain_perf}

\begin{table*}[!t]
\centering
\small{
\begin{tabular}{ccrrrrrrrrrrrrr}
\toprule
\multicolumn{2}{l}{} & \multicolumn{3}{c}{\phy} & \multicolumn{3}{c}{\chem} & \multicolumn{3}{c}{\bio} & \multicolumn{3}{c}{\overall} \\
\cmidrule(lr){3-5} \cmidrule(lr){6-8} \cmidrule(lr){9-11} \cmidrule(lr){12-14}
 & & \multicolumn{1}{c}{P} & \multicolumn{1}{c}{R} & \multicolumn{1}{c}{F} & \multicolumn{1}{c}{P} & \multicolumn{1}{c}{R} & \multicolumn{1}{c}{F} & \multicolumn{1}{c}{P} & \multicolumn{1}{c}{R} & \multicolumn{1}{c}{F} & \multicolumn{1}{c}{P} & \multicolumn{1}{c}{R} & \multicolumn{1}{c}{F} \\
\midrule
\multirow{2}{*}{\gfmini} & \dexpt & 0.56 & 0.45 & 0.38 & 0.65 & 0.55 & 0.49 & 0.60 & 0.47 & 0.42 & 0.60 & 0.48 & 0.43 \\
 & \rdexpt & 0.52 & 0.47 & 0.43 & 0.57 & 0.54 & 0.51 & 0.57 & 0.50 & 0.46 & 0.56 & 0.50 & 0.47 \\
\midrule
\multirow{2}{*}{\gfomni} & \dexpt & 0.53 & 0.47 & 0.43 & 0.69 & 0.59 & 0.57 & 0.73 & 0.54 & 0.52 & 0.68 & 0.54 & 0.51 \\
 & \rdexpt & 0.55 & 0.50 & 0.48 & 0.64 & 0.60 & 0.57 & 0.69 & 0.59 & 0.56 & 0.65 & 0.57 & 0.55 \\
\midrule
\multirow{2}{*}{\sonnet} & \dexpt & 0.62 & 0.53 & 0.50 & 0.68 & 0.62 & 0.61 & 0.76 & 0.68 & 0.68 & 0.71 & 0.65 & 0.64 \\
 & \rdexpt & 0.62 & 0.53 & 0.51 & 0.72 & 0.66 & 0.66 & 0.78 & 0.70 & 0.70 & 0.73 & 0.66 & 0.66 \\
\midrule
\othree & \rdexpt & 0.58 & 0.55 & 0.54 & 0.72 & 0.71 & 0.71 & 0.79 & 0.77 & 0.77 & 0.73 & 0.72 & 0.72 \\
\midrule
\ofour & \rdexpt & 0.64 & 0.59 & 0.57 & 0.73 & 0.72 & 0.71 & 0.81 & 0.77 & 0.77 & 0.75 & 0.72 & 0.72 \\
\midrule
\midrule
\multirow{2}{*}{\phimm} & \dexpt & 0.43 & 0.40 & 0.38 & 0.60 & 0.48 & 0.47 & 0.63 & 0.47 & 0.43 & 0.54 & 0.46 & 0.43 \\
 & \rdexpt & 0.62 & 0.37 & 0.27 & 0.52 & 0.39 & 0.32 & 0.60 & 0.42 & 0.37 & 0.58 & 0.41 & 0.34 \\
\midrule
\multirow{2}{*}{\llava} & \dexpt & 0.28 & 0.39 & 0.30 & 0.36 & 0.46 & 0.37 & 0.74 & 0.42 & 0.32 & 0.68 & 0.42 & 0.33 \\
 & \rdexpt & 0.52 & 0.39 & 0.34 & 0.55 & 0.45 & 0.40 & 0.50 & 0.44 & 0.38 & 0.51 & 0.43 & 0.38 \\
\midrule
\multirow{2}{*}{\llama} & \dexpt & 0.51 & 0.45 & 0.41 & 0.56 & 0.48 & 0.44 & 0.60 & 0.50 & 0.47 & 0.57 & 0.49 & 0.45 \\
 & \rdexpt & 0.53 & 0.41 & 0.35 & 0.52 & 0.41 & 0.34 & 0.60 & 0.43 & 0.36 & 0.56 & 0.42 & 0.35 \\
\midrule
\multirow{2}{*}{\molmo} & \dexpt & 0.46 & 0.41 & 0.37 & 0.55 & 0.47 & 0.42 & 0.62 & 0.50 & 0.46 & 0.57 & 0.47 & 0.43 \\
 & \rdexpt & 0.43 & 0.40 & 0.38 & 0.44 & 0.41 & 0.37 & 0.47 & 0.43 & 0.41 & 0.46 & 0.42 & 0.39 \\
\midrule
\multirow{2}{*}{\ivl} & \dexpt & 0.59 & 0.58 & 0.57 & 0.73 & 0.72 & 0.72 & 0.77 & 0.74 & 0.74 & 0.72 & 0.70 & 0.70 \\
 & \rdexpt & 0.56 & 0.49 & 0.46 & 0.69 & 0.57 & 0.56 & 0.75 & 0.59 & 0.58 & 0.69 & 0.57 & 0.55 \\
\midrule
\multirow{2}{*}{\qwen} & \dexpt & 0.68 & 0.58 & 0.54 & 0.70 & 0.65 & 0.62 & 0.72 & 0.66 & 0.65 & 0.70 & 0.65 & 0.63 \\
 & \rdexpt & 0.63 & 0.47 & 0.43 & 0.70 & 0.53 & 0.51 & 0.71 & 0.54 & 0.52 & 0.68 & 0.53 & 0.50 \\
\midrule
\multirow{2}{*}{\ds} & \dexpt & 0.47 & 0.47 & 0.46 & 0.52 & 0.50 & 0.49 & 0.51 & 0.51 & 0.50 & 0.50 & 0.67 & 0.58 \\
 & \rdexpt & 0.42 & 0.42 & 0.39 & 0.46 & 0.46 & 0.42 & 0.50 & 0.47 & 0.46 & 0.48 & 0.46 & 0.44 \\
\bottomrule
\end{tabular}
}
\caption{
Model performance on the claim verification task of \ourdata by the scientific domain of the claims when prompted to simply generate the decision (\dexpt), and when asked to reason and then generating the decision (\rdexpt).}
\label{tab:domain-results}
\end{table*}

\autoref{tab:domain-results} presents model performance for different domains in \ourdatasmall.
On average, we note that models are worst at verifying \phy claims and best at judging \chem claims.
However, the highest performance is achieved by \othree and \ofour on \bio claims.
We also present per-label metrics for model performance in each domain in \autoref{tab:phy-results}, \autoref{tab:chem-results} and \autoref{tab:bio-results}.

\begin{table*}[!t]
\centering
\small{
\begin{tabular}{ccrrrrrrrrrrrrr}
\toprule
\multicolumn{2}{l}{} & \multicolumn{3}{c}{\support} & \multicolumn{3}{c}{\neutral} & \multicolumn{3}{c}{\contradict} & \multicolumn{3}{c}{\overall} \\
\cmidrule(lr){3-5} \cmidrule(lr){6-8} \cmidrule(lr){9-11} \cmidrule(lr){12-14}
 & & \multicolumn{1}{c}{P} & \multicolumn{1}{c}{R} & \multicolumn{1}{c}{F} & \multicolumn{1}{c}{P} & \multicolumn{1}{c}{R} & \multicolumn{1}{c}{F} & \multicolumn{1}{c}{P} & \multicolumn{1}{c}{R} & \multicolumn{1}{c}{F} & \multicolumn{1}{c}{P} & \multicolumn{1}{c}{R} & \multicolumn{1}{c}{F} \\
\midrule
\multirow{2}{*}{\gfmini} & \dexpt & 0.40 & 0.81 & 0.54 & 0.52 & 0.47 & 0.49 & 0.75 & 0.06 & 0.12 & 0.56 & 0.45 & 0.38 \\
 & \rdexpt & 0.41 & 0.79 & 0.54 & 0.56 & 0.46 & 0.50 & 0.58 & 0.15 & 0.23 & 0.52 & 0.47 & 0.43 \\
\midrule
\multirow{2}{*}{\gfomni} & \dexpt & 0.40 & 0.83 & 0.54 & 0.63 & 0.38 & 0.47 & 0.56 & 0.19 & 0.28 & 0.53 & 0.47 & 0.43 \\
 & \rdexpt & 0.44 & 0.71 & 0.54 & 0.54 & 0.57 & 0.56 & 0.66 & 0.22 & 0.33 & 0.55 & 0.50 & 0.48 \\
\midrule
\multirow{2}{*}{\sonnet} & \dexpt & 0.44 & 0.77 & 0.56 & 0.60 & 0.58 & 0.59 & 0.81 & 0.23 & 0.36 & 0.62 & 0.53 & 0.50 \\
 & \rdexpt & 0.45 & 0.79 & 0.57 & 0.60 & 0.54 & 0.57 & 0.81 & 0.26 & 0.39 & 0.62 & 0.53 & 0.51 \\
\midrule
\othree & \rdexpt & \textbf{0.59} & 0.52 & 0.55 & 0.50 & 0.79 & 0.62 & 0.63 & \textbf{0.34} & 0.45 & 0.58 & 0.55 & 0.54 \\
\midrule
\ofour & \rdexpt & 0.53 & 0.66 & 0.59 & 0.56 & 0.76 & 0.65 & 0.82 & \textbf{0.34} & \textbf{0.49} & 0.64 & \textbf{0.59} & \textbf{0.57} \\
\midrule
\midrule
\multirow{2}{*}{\phimm} & \dexpt & 0.37 & 0.70 & 0.49 & 0.43 & 0.24 & 0.31 & 0.48 & 0.27 & 0.35 & 0.43 & 0.40 & 0.38 \\
 & \rdexpt & 0.35 & \textbf{0.90} & 0.50 & \textbf{1.00} & 0.02 & 0.04 & 0.51 & 0.19 & 0.27 & 0.62 & 0.37 & 0.27 \\
\midrule
\multirow{2}{*}{\llava} & \dexpt & 0.36 & 0.78 & 0.49 & 0.47 & 0.39 & 0.42 & 0.00 & 0.00 & 0.00 & 0.28 & 0.39 & 0.30 \\
 & \rdexpt & 0.37 & 0.74 & 0.49 & 0.40 & 0.35 & 0.38 & 0.80 & 0.08 & 0.15 & 0.52 & 0.39 & 0.34 \\
\midrule
\multirow{2}{*}{\llama} & \dexpt & 0.40 & 0.80 & 0.53 & 0.54 & 0.41 & 0.46 & 0.59 & 0.14 & 0.22 & 0.51 & 0.45 & 0.41 \\
 & \rdexpt & 0.36 & \textbf{0.90} & 0.52 & 0.59 & 0.18 & 0.27 & 0.65 & 0.16 & 0.25 & 0.53 & 0.41 & 0.35 \\
\midrule
\multirow{2}{*}{\molmo} & \dexpt & 0.38 & 0.83 & 0.52 & 0.51 & 0.22 & 0.31 & 0.50 & 0.19 & 0.27 & 0.46 & 0.41 & 0.37 \\
 & \rdexpt & 0.37 & 0.72 & 0.49 & 0.50 & 0.23 & 0.31 & 0.43 & 0.26 & 0.32 & 0.43 & 0.40 & 0.38 \\
\midrule
\multirow{2}{*}{\ivl} & \dexpt & 0.55 & 0.69 & \textbf{0.61} & 0.59 & 0.68 & 0.63 & 0.63 & 0.38 & 0.47 & 0.59 & 0.58 & \textbf{0.57} \\
 & \rdexpt & 0.43 & 0.78 & 0.55 & 0.55 & 0.49 & 0.52 & 0.70 & 0.20 & 0.31 & 0.56 & 0.49 & 0.46 \\
\midrule
\multirow{2}{*}{\qwen} & \dexpt & 0.52 & 0.67 & 0.59 & 0.56 & \textbf{0.82} & \textbf{0.66} & \textbf{0.96} & 0.24 & 0.38 & \textbf{0.68} & 0.58 & 0.54 \\
 & \rdexpt & 0.40 & 0.84 & 0.54 & 0.55 & 0.39 & 0.45 & 0.94 & 0.18 & 0.30 & 0.63 & 0.47 & 0.43 \\
\midrule
\multirow{2}{*}{\ds} & \dexpt & 0.52 & 0.43 & 0.47 & 0.48 & 0.72 & 0.57 & 0.41 & 0.27 & 0.33 & 0.47 & 0.47 & 0.46 \\
 & \rdexpt & 0.40 & 0.60 & 0.48 & 0.45 & 0.50 & 0.47 & 0.43 & 0.16 & 0.23 & 0.42 & 0.42 & 0.39 \\
\bottomrule
\end{tabular}
}
\caption{
Model performance on the claim verification task of \ourdata for \phy claims when prompted to simply generate the decision (\dexpt), and when asked to reason and then generating the decision (\rdexpt).}
\label{tab:phy-results}
\end{table*}

\begin{table*}[!t]
\centering
\small{
\begin{tabular}{ccrrrrrrrrrrrrr}
\toprule
\multicolumn{2}{l}{} & \multicolumn{3}{c}{\support} & \multicolumn{3}{c}{\neutral} & \multicolumn{3}{c}{\contradict} & \multicolumn{3}{c}{\overall} \\
\cmidrule(lr){3-5} \cmidrule(lr){6-8} \cmidrule(lr){9-11} \cmidrule(lr){12-14}
 & & \multicolumn{1}{c}{P} & \multicolumn{1}{c}{R} & \multicolumn{1}{c}{F} & \multicolumn{1}{c}{P} & \multicolumn{1}{c}{R} & \multicolumn{1}{c}{F} & \multicolumn{1}{c}{P} & \multicolumn{1}{c}{R} & \multicolumn{1}{c}{F} & \multicolumn{1}{c}{P} & \multicolumn{1}{c}{R} & \multicolumn{1}{c}{F} \\
\midrule
\multirow{2}{*}{\gfmini} & \dexpt & 0.46 & 0.86 & 0.60 & 0.70 & 0.69 & 0.69 & 0.79 & 0.11 & 0.19 & 0.65 & 0.55 & 0.49 \\
 & \rdexpt & 0.46 & 0.74 & 0.57 & 0.63 & 0.67 & 0.65 & 0.63 & 0.21 & 0.32 & 0.57 & 0.54 & 0.51 \\
\midrule
\multirow{2}{*}{\gfomni} & \dexpt & 0.46 & 0.91 & 0.61 & 0.91 & 0.59 & 0.72 & 0.70 & 0.25 & 0.37 & 0.69 & 0.59 & 0.57 \\
 & \rdexpt & 0.49 & 0.82 & 0.61 & 0.76 & 0.71 & 0.73 & 0.66 & 0.26 & 0.38 & 0.64 & 0.60 & 0.57 \\
\midrule
\multirow{2}{*}{\sonnet} & \dexpt & 0.49 & 0.85 & 0.63 & 0.89 & 0.62 & 0.73 & 0.66 & 0.38 & 0.48 & 0.68 & 0.62 & 0.61 \\
 & \rdexpt & 0.53 & 0.87 & 0.66 & 0.90 & 0.63 & 0.74 & 0.74 & 0.48 & 0.58 & 0.72 & 0.66 & 0.66 \\
\midrule
\othree & \rdexpt & \textbf{0.75} & 0.66 & 0.70 & 0.67 & 0.86 & 0.76 & 0.73 & 0.61 & \textbf{0.67} & 0.72 & 0.71 & 0.71 \\
\midrule
\ofour & \rdexpt & 0.67 & 0.78 & \textbf{0.72} & 0.72 & 0.82 & 0.76 & 0.81 & 0.56 & 0.66 & \textbf{0.73} & \textbf{0.72} & 0.71 \\
\midrule
\midrule
\multirow{2}{*}{\phimm} & \dexpt & 0.43 & 0.71 & 0.54 & \textbf{0.97} & 0.29 & 0.45 & 0.41 & 0.44 & 0.42 & 0.60 & 0.48 & 0.47 \\
 & \rdexpt & 0.35 & 0.86 & 0.50 & 0.56 & 0.05 & 0.09 & 0.65 & 0.27 & 0.38 & 0.52 & 0.39 & 0.32 \\
\midrule
\multirow{2}{*}{\llava} & \dexpt & 0.39 & 0.90 & 0.55 & 0.68 & 0.47 & 0.55 & 0.00 & 0.00 & 0.00 & 0.36 & 0.46 & 0.37 \\
 & \rdexpt & 0.38 & 0.80 & 0.51 & 0.60 & 0.45 & 0.51 & 0.69 & 0.11 & 0.18 & 0.55 & 0.45 & 0.40 \\
\midrule
\multirow{2}{*}{\llama} & \dexpt & 0.40 & 0.86 & 0.55 & 0.81 & 0.38 & 0.52 & 0.46 & 0.18 & 0.26 & 0.56 & 0.48 & 0.44 \\
 & \rdexpt & 0.38 & 0.93 & 0.54 & 0.67 & 0.12 & 0.20 & 0.53 & 0.18 & 0.27 & 0.52 & 0.41 & 0.34 \\
\midrule
\multirow{2}{*}{\molmo} & \dexpt & 0.42 & \textbf{0.92} & 0.58 & 0.78 & 0.27 & 0.40 & 0.44 & 0.20 & 0.28 & 0.55 & 0.47 & 0.42 \\
 & \rdexpt & 0.38 & 0.80 & 0.52 & 0.58 & 0.28 & 0.38 & 0.36 & 0.16 & 0.22 & 0.44 & 0.41 & 0.37 \\
\midrule
\multirow{2}{*}{\ivl} & \dexpt & 0.66 & 0.76 & 0.70 & 0.88 & 0.74 & 0.80 & 0.65 & \textbf{0.66} & 0.66 & \textbf{0.73} & \textbf{0.72} & \textbf{0.72} \\
 & \rdexpt & 0.44 & 0.89 & 0.59 & 0.96 & 0.53 & 0.69 & 0.66 & 0.28 & 0.39 & 0.69 & 0.57 & 0.56 \\
\midrule
\multirow{2}{*}{\qwen} & \dexpt & 0.56 & 0.74 & 0.64 & 0.68 & \textbf{0.90} & \textbf{0.78} & \textbf{0.86} & 0.31 & 0.46 & 0.70 & 0.65 & 0.62 \\
 & \rdexpt & 0.41 & 0.91 & 0.57 & 0.88 & 0.43 & 0.58 & 0.81 & 0.24 & 0.37 & 0.70 & 0.53 & 0.51 \\
\midrule
\multirow{2}{*}{\ds} & \dexpt & 0.62 & 0.36 & 0.45 & 0.50 & 0.76 & 0.60 & 0.43 & 0.39 & 0.41 & 0.52 & 0.50 & 0.49 \\
 & \rdexpt & 0.41 & 0.69 & 0.52 & 0.53 & 0.54 & 0.54 & 0.44 & 0.14 & 0.21 & 0.46 & 0.46 & 0.42 \\
\bottomrule
\end{tabular}
}
\caption{
Model performance on the claim verification task of \ourdata for \chem claims when prompted to simply generate the decision (\dexpt), and when asked to reason and then generating the decision (\rdexpt).}
\label{tab:chem-results}
\end{table*}

\begin{table*}[!t]
\centering
\small{
\begin{tabular}{ccrrrrrrrrrrrrr}
\toprule
\multicolumn{2}{l}{} & \multicolumn{3}{c}{\support} & \multicolumn{3}{c}{\neutral} & \multicolumn{3}{c}{\contradict} & \multicolumn{3}{c}{\overall} \\
\cmidrule(lr){3-5} \cmidrule(lr){6-8} \cmidrule(lr){9-11} \cmidrule(lr){12-14}
 & & \multicolumn{1}{c}{P} & \multicolumn{1}{c}{R} & \multicolumn{1}{c}{F} & \multicolumn{1}{c}{P} & \multicolumn{1}{c}{R} & \multicolumn{1}{c}{F} & \multicolumn{1}{c}{P} & \multicolumn{1}{c}{R} & \multicolumn{1}{c}{F} & \multicolumn{1}{c}{P} & \multicolumn{1}{c}{R} & \multicolumn{1}{c}{F} \\
\midrule
\multirow{2}{*}{\gfmini} & \dexpt & 0.40 & 0.90 & 0.56 & 0.66 & 0.42 & 0.51 & 0.74 & 0.10 & 0.18 & 0.60 & 0.47 & 0.42 \\
 & \rdexpt & 0.42 & 0.87 & 0.57 & 0.68 & 0.41 & 0.51 & 0.62 & 0.21 & 0.31 & 0.57 & 0.50 & 0.46 \\
\midrule
\multirow{2}{*}{\gfomni} & \dexpt & 0.43 & 0.96 & 0.59 & 0.93 & 0.44 & 0.60 & 0.84 & 0.23 & 0.36 & 0.73 & 0.54 & 0.52 \\
 & \rdexpt & 0.48 & 0.92 & 0.63 & 0.76 & 0.58 & 0.66 & 0.83 & 0.26 & 0.39 & 0.69 & 0.59 & 0.56 \\
\midrule
\multirow{2}{*}{\sonnet} & \dexpt & 0.53 & 0.92 & 0.67 & 0.91 & 0.63 & 0.74 & 0.82 & 0.49 & 0.61 & 0.76 & 0.68 & 0.68 \\
 & \rdexpt & 0.55 & 0.92 & 0.69 & 0.94 & 0.64 & 0.76 & 0.84 & 0.54 & 0.65 & 0.78 & 0.70 & 0.70 \\
\midrule
\othree & \rdexpt & \textbf{0.67} & 0.84 & \textbf{0.75} & 0.80 & \textbf{0.77} & 0.78 & 0.89 & 0.69 & \textbf{0.78} & 0.79 & \textbf{0.77} & \textbf{0.77} \\
\midrule
\ofour & \rdexpt & 0.63 & 0.92 & \textbf{0.75} & 0.87 & 0.75 & \textbf{0.80} & 0.92 & 0.64 & 0.76 & \textbf{0.81} & \textbf{0.77} & \textbf{0.77} \\
\midrule
\midrule
\multirow{2}{*}{\phimm} & \dexpt & 0.46 & 0.69 & 0.55 & \textbf{0.98} & 0.14 & 0.24 & 0.44 & 0.60 & 0.51 & 0.63 & 0.47 & 0.43 \\
 & \rdexpt & 0.36 & 0.86 & 0.51 & 0.85 & 0.13 & 0.23 & 0.57 & 0.27 & 0.37 & 0.60 & 0.42 & 0.37 \\
\midrule
\multirow{2}{*}{\llava} & \dexpt & 0.37 & \textbf{0.98} & 0.53 & 0.85 & 0.27 & 0.41 & \textbf{1.00} & 0.01 & 0.01 & 0.74 & 0.42 & 0.32 \\
 & \rdexpt & 0.39 & 0.82 & 0.53 & 0.58 & 0.42 & 0.49 & 0.55 & 0.08 & 0.14 & 0.50 & 0.44 & 0.38 \\
\midrule
\multirow{2}{*}{\llama} & \dexpt & 0.42 & 0.88 & 0.57 & 0.70 & 0.44 & 0.54 & 0.67 & 0.18 & 0.29 & 0.60 & 0.50 & 0.47 \\
 & \rdexpt & 0.38 & 0.95 & 0.54 & 0.79 & 0.16 & 0.27 & 0.63 & 0.17 & 0.27 & 0.60 & 0.43 & 0.36 \\
\midrule
\multirow{2}{*}{\molmo} & \dexpt & 0.42 & 0.92 & 0.57 & 0.85 & 0.32 & 0.47 & 0.60 & 0.24 & 0.34 & 0.62 & 0.50 & 0.46 \\
 & \rdexpt & 0.39 & 0.75 & 0.51 & 0.57 & 0.30 & 0.39 & 0.44 & 0.25 & 0.32 & 0.47 & 0.43 & 0.41 \\
\midrule
\multirow{2}{*}{\ivl} & \dexpt & 0.63 & 0.84 & 0.72 & 0.93 & 0.66 & 0.77 & 0.74 & \textbf{0.71} & 0.72 & 0.77 & 0.74 & \textbf{0.74} \\
 & \rdexpt & 0.46 & 0.96 & 0.62 & 0.93 & 0.47 & 0.62 & 0.88 & 0.35 & 0.50 & 0.75 & 0.59 & 0.58 \\
\midrule
\multirow{2}{*}{\qwen} & \dexpt & 0.55 & 0.85 & 0.67 & 0.77 & 0.75 & 0.76 & 0.83 & 0.37 & 0.51 & 0.72 & 0.66 & 0.65 \\
 & \rdexpt & 0.43 & 0.93 & 0.59 & 0.86 & 0.42 & 0.56 & 0.84 & 0.28 & 0.43 & 0.71 & 0.54 & 0.52 \\
\midrule
\multirow{2}{*}{\ds} & \dexpt & 0.55 & 0.45 & 0.49 & 0.52 & 0.63 & 0.57 & 0.46 & 0.44 & 0.45 & 0.51 & 0.51 & 0.50 \\
 & \rdexpt & 0.42 & 0.66 & 0.51 & 0.53 & 0.50 & 0.51 & 0.54 & 0.26 & 0.35 & 0.50 & 0.47 & 0.46 \\
\bottomrule
\end{tabular}
}
\caption{
Model performance on the claim verification task of \ourdata for \bio claims when prompted to simply generate the decision (\dexpt), and when asked to reason and then generating the decision (\rdexpt).}
\label{tab:bio-results}
\end{table*}

\begin{table*}[!t]
\centering
\begin{tabular}{ccrrrrrrrrrr}
\toprule
\multicolumn{2}{l}{} & \multicolumn{3}{c}{\support} & \multicolumn{3}{c}{\nsup} & \multicolumn{3}{c}{\overall} \\
\cmidrule(lr){3-5} \cmidrule(lr){6-8} \cmidrule(lr){9-11}
 &  & P & R & F & P & R & F & P & R & F \\
\hline
\multirow{3}{*}{\gfmini} & D & 0.39 & 0.93 & 0.55 & 0.89 & 0.29 & 0.43 & 0.72 & 0.5 & 0.47 \\
 & R$\rightarrow$D & 0.41 & 0.92 & 0.56 & 0.89 & 0.33 & 0.49 & 0.73 & 0.53 & 0.51 \\
 & I$\rightarrow$R$\rightarrow$D & 0.37 & 0.96 & 0.53 & 0.91 & 0.18 & 0.30 & 0.73 & 0.44 & 0.37 \\
\hline
\multirow{3}{*}{\gfomni} & D & 0.41 & 0.96 & 0.57 & 0.94 & 0.30 & 0.46 & 0.77 & 0.52 & 0.50 \\
 & R$\rightarrow$D & 0.43 & 0.95 & 0.59 & 0.94 & 0.38 & 0.54 & 0.77 & 0.57 & 0.56 \\
 & I$\rightarrow$R$\rightarrow$D & 0.39 & 0.98 & 0.56 & 0.97 & 0.25 & 0.39 & 0.78 & 0.49 & 0.45 \\
\midrule
\multirow{3}{*}{\sonnet} & D & 0.52 & 0.93 & 0.67 & 0.94 & 0.57 & 0.71 & 0.80 & 0.69 & 0.70 \\
 & R$\rightarrow$D & 0.53 & 0.95 & 0.68 & 0.96 & 0.58 & 0.72 & 0.82 & 0.70 & 0.71 \\
 & I$\rightarrow$R$\rightarrow$D & 0.51 & 0.96 & 0.66 & 0.96 & 0.53 & 0.69 & 0.81 & 0.68 & 0.68 \\
\bottomrule
\end{tabular}
\caption{Model performance on \bio claims when posing \ourdatasmall as a two class problem.}
\label{tab:app-2class}
\end{table*}

\subsection{Few-Shot Experiments}

\begin{table*}
\small
\centering
\begin{tabular}{ l | c | c | c | c | c | c | c | c | c | c | c | c | c }
\toprule
\multirow{2}{*}{} & \multirow{2}{*}{\# Examples} & \multicolumn{3}{c|}{\support} & \multicolumn{3}{c|}{\neutral} & \multicolumn{3}{c|}{\contradict} & \multicolumn{3}{c}{\overall} \\
 &  & P & R & F & P & R & F & P & R & F & P & R & F \\
\midrule
\multirow{4}{*}{4o-mini} & $k=0$ & 0.40 & 0.90 & 0.56 & 0.66 & 0.42 & 0.51 & 0.74 & 0.10 & 0.18 & 0.6 & 0.47 & 0.42 \\
 & $k=1$ & 0.41 & 0.90 & 0.56 & 0.81 & 0.12 & 0.22 & 0.73 & 0.48 & 0.58 & 0.65 & 0.50 & 0.45 \\
 & $k=3$ & 0.41 & 0.92 & 0.57 & 0.71 & 0.46 & 0.56 & 0.8 & 0.11 & 0.19 & 0.64 & 0.50 & 0.44 \\
 & $k=5$ & 0.41 & 0.91 & 0.56 & 0.71 & 0.49 & 0.58 & 0.86 & 0.08 & 0.15 & 0.66 & 0.49 & 0.43 \\
\midrule
\multirow{4}{*}{GPT-4o} & $k=0$ & 0.43 & 0.96 & 0.59 & 0.93 & 0.44 & 0.60 & 0.84 & 0.23 & 0.36 & 0.73 & 0.54 & 0.52 \\
 & $k=1$ & 0.48 & 0.92 & 0.63 & 0.90 & 0.53 & 0.66 & 0.77 & 0.38 & 0.51 & 0.72 & 0.61 & 0.60 \\
 & $k=3$ & 0.46 & 0.95 & 0.62 & 0.90 & 0.53 & 0.67 & 0.83 & 0.28 & 0.42 & 0.73 & 0.59 & 0.57 \\
 & $k=5$ & 0.45 & 0.95 & 0.62 & 0.90 & 0.51 & 0.65 & 0.86 & 0.28 & 0.42 & 0.74 & 0.58 & 0.56 \\
\midrule
\multirow{4}{*}{Sonnet} & $k=0$ & 0.53 & 0.92 & 0.67 & 0.91 & 0.63 & 0.74 & 0.82 & 0.49 & 0.61 & 0.76 & 0.68 & 0.68 \\
 & $k=1$ & 0.64 & 0.73 & 0.68 & 0.74 & 0.74 & 0.74 & 0.76 & 0.66 & 0.71 & 0.71 & 0.71 & 0.71 \\
 & $k=3$ & 0.62 & 0.75 & 0.68 & 0.75 & 0.66 & 0.70 & 0.73 & 0.65 & 0.69 & 0.70 & 0.69 & 0.69 \\
 & $k=5$ & 0.60 & 0.83 & 0.69 & 0.83 & 0.69 & 0.76 & 0.77 & 0.60 & 0.67 & 0.73 & 0.71 & 0.71 \\
 \bottomrule
\end{tabular}
\caption{Model (\dexpt) performance on \bio claims in \ourdatasmall. $k$ denotes the number of few-shot examples provided as part of the prompt.}
\label{tab:fewshot}
\end{table*}

We experiment with few-shot prompting (or in-context learning) \cite{brown-etal-2020-fewshot, incontext} on \bio claims for the decision-only (\dexpt) experiments using a subset of closed-source models.
Using the methodology described in \autoref{sec:data_creation}, we create 45 claims from Cell papers \footnote{These papers are not part of the \ourdatasmall evaluation set.}.
The few-shot examples ($k$) are selected randomly from these created claims.

\autoref{tab:fewshot} presents model performance when prompted to just produce a decision (\dexpt).
We find that performance improves for all models when they are provided any number of in-context examples.
However, the benefits go down as the number of examples goes up.

\subsection{\ourdata task as a two-class problem} 

\autoref{tab:app-2class} presents the results when the main claim verification task of \ourdata is converted from a three-class (\support, \contradict, \neutral) problem to a two-class problem by merging \contradict and \neutral classes to \nsup class. From the table, we observe that overall F1-scores do vary from the three-class F1-scores, highlighting that \ourdata is hard to solve even on a simplified problem setting. We see higher precision and recall values for \nsup compared \contradict and \neutral metrics in three-class problem. This shows that while models can do coarse-grained classification of wrong or irrelevant claims, in context of the figure and cpation, but struggle when doing fine-grained classification.

\subsection{Panel Complexity}

\autoref{tab:app-complexity} shows the results of different models when doing inference on single panel images from \ourdata compared to multi-panel images. Multi-panel images represent claim verification tasks from \ourdata of higher complexity since models have to reason on the correct panel and filter out distractor panels. However, results show models doing better on average for multi-panel setting compared to single-panel setting. This might be because multi-panel provides more visual context for model to do the task.

\begin{table*}[!t]
\centering
\begin{tabular}{c | c | r | r | r | r | r | r }
\hline
\multicolumn{2}{l|}{\multirow{2}{*}{}} & \multicolumn{3}{c}{Single-Panel} & \multicolumn{3}{|c}{Multi-Panel} \\
\hline
 &  & P & R & F & P & R & F \\
\hline
\multirow{3}{*}{\gfmini} & D & 0.59 & 0.47 & 0.41 & 0.65 & 0.51 & 0.47 \\
 & R$\rightarrow$D & 0.58 & 0.49 & 0.46 & 0.56 & 0.51 & 0.48 \\
 & I$\rightarrow$R$\rightarrow$D & 0.6 & 0.45 & 0.4 & 0.57 & 0.46 & 0.41 \\
\hline
\multirow{3}{*}{\gfomni} & D & 0.73 & 0.54 & 0.52 & 0.77 & 0.55 & 0.53 \\
 & R$\rightarrow$D & 0.69 & 0.58 & 0.56 & 0.72 & 0.61 & 0.59 \\
 & I$\rightarrow$R$\rightarrow$D & 0.73 & 0.51 & 0.47 & 0.76 & 0.53 & 0.5 \\
\hline
\multirow{3}{*}{\sonnet} & D & 0.75 & 0.68 & 0.68 & 0.78 & 0.68 & 0.69 \\
 & R$\rightarrow$D & 0.78 & 0.7 & 0.7 & 0.79 & 0.72 & 0.72 \\
 & I$\rightarrow$R$\rightarrow$D & 0.79 & 0.7 & 0.7 & 0.76 & 0.67 & 0.68 \\
\hline
\end{tabular}
\caption{Model performance on \bio claims when broken down by complexity of visual aggregation for claims}
\label{tab:app-complexity}
\end{table*}

\section{Qualitative Error Analysis}
\label{appsec:error-analysis}

In addition to the diagnostics, we perform qualitative error analysis on a random sample of 100 errors in \othree reasoning (\rdexpt) on \ourdatasmall.
We categorize the errors into the following categories:
\squishlist
\item[1.] Domain Expertise (27\%) - Models lack domain expertise, knowledge of related work and common practices of representing data specific to the scientific field.
\item[2.] Visual Understanding (23\%) - Models are unable to make the correct inference from the information that they perceive from the figure.
\item[3.] Visual Perception (23\%) - Models either miss or pick up data presented in the figure, which can lead to the correct judgment.
\item[4.] Cross Modal Aggregation (17\%) - Models incorrectly weight information from one modality over another (such as focus on the caption more than the figure), coming to the wrong conclusion.
\item[5.] Others (10\%) - This category contains infrequent error types bucketed together. For instance, models simply misunderstand the caption (textual understanding), or fail to aggregate information from multiple panels (multi-panel aggregation).
\squishend

\section{Prompts Used}
\label{appsec:prompts}

We present the exact prompts used for different experiments with \sonnet in \autoref{fig:closed-model-decision-prompt}, \autoref{fig:closed-model-reasoning-decision-prompt} and \autoref{fig:closed-model-identify-reasoning-decision-prompt} and \ivl in \autoref{fig:open-model-decision-prompt}, \autoref{fig:open-model-reasoning-decision-prompt} and \autoref{fig:open-model-identify-reasoning-decision-prompt}.

\begin{figure}[!t]
\small
\centerline{
\fbox{
    \parbox{0.95\columnwidth}{
    You are an AI model tasked with verifying claims related to visual evidence using zero-shot learning. Your job is to analyze a given image(s) and its provided caption(s) to decide whether it SUPPORT or CONTRADICT or NEUTRAL the provided claim.\\

    CLAIM: {CLAIM}\\
    IMAGE CAPTION(S): {IMAGE\_CAPTIONS}\\

    Guidelines:\\
    1. Evaluate the claim’s plausibility based on visual elements within the image(s).\\
    2. Consider the relevance, meaning, and implications of both the depicted content and the caption(s).\\
    3. Analyze the broader context and scope of the image(s) and caption(s) in relation to the claim.\\

    After completing your analysis, output exactly one JSON object with exactly one key: ``decision''.\\
    - For ``decision'', output exactly one word — either ``SUPPORT'' or ``CONTRADICT'' or ``NEUTRAL'' (uppercase, no extra text).\\
    Do NOT add markdown formatting, code fences, or any additional text. The output must start with an opening curly brace \{ and end with a closing curly brace \}.\\

    Example output format:\\
    \{``decision'': ``SUPPORT''\}\\

    Now, please evaluate the image(s) and caption(s) with respect to the claim provided above.
    }
}}
\caption{Prompt for \sonnet for the \dexpt experiment}
\label{fig:closed-model-decision-prompt}
\end{figure}

\begin{figure}[!t]
\small
\centerline{
\fbox{
    \parbox{0.95\columnwidth}{
    You are an AI model tasked with verifying claims related to visual evidence using zero-shot learning. Your job is to analyze a given image(s) and its provided caption(s) to decide whether it SUPPORT or CONTRADICT or NEUTRAL the provided claim.\\
    CLAIM: \{CLAIM\}\\
    IMAGE CAPTION(S): \{IMAGE\_CAPTIONS\}\\
    Guidelines:\\
    1. Evaluate the claim’s plausibility based on visual elements within the image(s). 2. Consider the relevance, meaning, and implications of both the depicted content and the caption(s). 3. Analyze the broader context and scope of the image(s) and caption(s) in relation to the claim. 4. Think step by step to reach your conclusion, but only provide a concise reasoning statement in the output.\\
    After completing your analysis, output exactly one JSON object with exactly two keys in this order: ``reasoning" and ``decision".\\
    - For ``reasoning", provide a brief (one- or two-sentence) explanation of your analysis.\\
    - For ``decision", output exactly one word — either ``SUPPORT" or ``CONTRADICT" or ``NEUTRAL" (uppercase, no extra text).\\
    Do NOT add markdown formatting, code fences, or any additional text. The output must start with an opening curly brace \{ and end with a closing curly brace \}.\\
    Example output format:\\
    \{``reasoning": ``The caption confirms the rising trend visible in the image, supporting the claim.", ``decision": ``SUPPORT"\}\\
    Now, please evaluate the image(s) and caption(s) with respect to the claim provided above.
    }
}}
\caption{Prompt for \sonnet for the \rdexpt experiment}
\label{fig:closed-model-reasoning-decision-prompt}
\end{figure}

\begin{figure}[!t]
\small
\centerline{
\fbox{
    \parbox{0.95\columnwidth}{
    You are an AI model tasked with verifying claims related to visual evidence using zero-shot learning. Your job is to analyze a given image(s) and its provided caption(s) to decide whether it SUPPORT or CONTRADICT or NEUTRAL the provided claim.\\

    CLAIM: {CLAIM}\\
    IMAGE CAPTION(S): {IMAGE\_CAPTIONS}\\

    Guidelines:\\
    1. Evaluate the claim’s plausibility based on visual elements within the image(s).\\
    2. Consider the relevance, meaning, and implications of both the depicted content and the caption(s).\\
    3. Analyze the broader context and scope of the image(s) and caption(s) in relation to the claim.\\
    4. Identify which specific panels (e.g., Panel A, Panel B, Panel C, etc.) are necessary to evaluate the claim.\\
    5. Think step by step to reach your conclusion and provide it in a concise manner in the output.\\

    After completing your analysis, output exactly one JSON object with exactly three keys in this order: ``figure\_panels'', ``reasoning'', and ``decision''.\\
    - For ``figure\_panels'', list ONLY the names or labels of the panels needed to evaluate the claim (e.g., [``Panel A'', ``Panel C'']) with no further description. If no panels are needed, return [].\\
    - For ``reasoning'', provide a brief (one- or two-sentence) explanation of your analysis.\\
    - For ``decision'', output exactly one word — either ``SUPPORT'' or ``CONTRADICT'' or ``NEUTRAL'' (uppercase, no extra text).\\
    Do NOT add markdown formatting, code fences, or any additional text. The output must start with an opening curly brace \{{ and end with a closing curly brace \}}.\\

    Example output format:\\
    \{{``figure\_panels'': [``Panel A'', ``Panel C''], ``reasoning``: ``The trend in Panel A aligns with the claim, while Panel C corroborates the effect.'', ``decision'': ``SUPPORT''\}}\\

    Now, please evaluate the image(s) and caption(s) with respect to the claim provided above.
    }
}}
\caption{Prompt for \sonnet for the \irdexpt experiment}
\label{fig:closed-model-identify-reasoning-decision-prompt}
\end{figure}

\begin{figure}[!t]
\small
\centerline{
\fbox{
    \parbox{0.95\columnwidth}{
This is an image from a scientific paper.
The following is the caption of the image.\\

IMAGE CAPTION(S): {IMAGE\_CAPTIONS}\\

Using this image, analyze whether the following claim is supported, contradicted or neutral according to the image and caption.\\

CLAIM: {CLAIM}\\

Reply with one of the following keywords: SUPPORT, CONTRADICT, NEUTRAL.
Do not generate any other text or explanation.\\

Return your answer in following format:\\
DECISION: <your decision>\\

    }
}}
\caption{Prompt for \ivl for the \dexpt experiment}
\label{fig:open-model-decision-prompt}
\end{figure}

\begin{figure}[!t]
\small
\centerline{
\fbox{
    \parbox{0.95\columnwidth}{
This is an image from a scientific paper. The following is the caption of this image.\\

IMAGE CAPTION(S): {IMAGE\_CAPTIONS}\\

Using this image, analyze whether the following claim is supported, contradicted or neutral according to the image and caption.\\

CLAIM: {CLAIM}\\

Think step by step to reach your conclusion and then reply with only one of the following keywords: SUPPORT, CONTRADICT, NEUTRAL. Your reasoning should be brief and concise, no more than 100 words.\\

Return your answer in following format:\\
REASONING: <your reasoning>\\
DECISION: <your decision>\\
    }
}}
\caption{Prompt for \ivl for the \rdexpt experiment}
\label{fig:open-model-reasoning-decision-prompt}
\end{figure}

\begin{figure}[!t]
\small
\centerline{
\fbox{
    \parbox{0.95\columnwidth}{
This is an image, with multiple panels, from a scientific paper. The following is the caption of this image.\\

IMAGE CAPTION(S): {IMAGE\_CAPTIONS}\\

Using this image, analyze whether the following claim is supported, contradicted or neutral according to the image and caption.\\

CLAIM: {CLAIM}\\ 

First identify the relevant panels (Figure A, Figure B etc.) in the image that are needed to analyze the claim. Then think step by step to reach your conclusion and reply with only one of the following keywords: SUPPORT, CONTRADICT, NEUTRAL. Your reasoning should be brief and concise, no more than 100 words.\\

Return your answer in following format:\\
FIGURE PANELS: <the figure panels to use for deduction>\\
REASONING: <your reasoning>\\
DECISION: <your decision>\\
    }
}}
\caption{Prompt for \ivl for the \irdexpt experiment}
\label{fig:open-model-identify-reasoning-decision-prompt}
\end{figure}

\begin{table*}
\centering
\small
\begin{tabular}{| c | c | c | c | c | c | c |}
\toprule
Dataset & Scientific? & Multimodal? & Claim Verification? & Heterogeneous Figures? & Real? & Complex? \\
\midrule
ArXivQa & \checkmark & \checkmark & \xmark & \checkmark & \checkmark & \xmark \\
MMC & \checkmark & \checkmark & \xmark & \checkmark & \checkmark & \xmark \\
PlotQA & \xmark & \checkmark & \xmark & \xmark & \xmark & \xmark \\
SPIQA & \checkmark & \checkmark & \xmark & \checkmark & \checkmark & \xmark \\
FigureQA & \xmark & \checkmark & \xmark & \xmark & \xmark & \xmark \\
DVQA & \xmark & \checkmark & \xmark & \xmark & \xmark & \xmark \\
ChartQA & \xmark & \checkmark & \xmark & \xmark & \checkmark & \xmark \\
ChartBench & \xmark & \checkmark & \xmark & \checkmark & \checkmark & \xmark \\
ChartX & \xmark & \checkmark & \xmark & \checkmark & \xmark & \xmark \\
MultiChartQA & \xmark & \checkmark & \xmark & \checkmark & \checkmark & \checkmark \\
ChartCheck & \xmark & \checkmark & \checkmark & \xmark & \checkmark & \xmark \\
SciFact & \checkmark & \xmark & \checkmark & \xmark & \xmark & \xmark \\
SciFact-Open & \checkmark & \xmark & \checkmark & \xmark & \xmark & \xmark \\
PubHealthTab & \xmark & \xmark & \checkmark & \xmark & \xmark & \xmark \\
MuSciClaims & \checkmark & \checkmark & \checkmark & \checkmark & \checkmark & \checkmark \\
\bottomrule
\end{tabular}
\caption{Comparison of \ourdata against related work benchmarks across different desired characteristics for a multimodal claim verification task.}
\label{tab:dataset_characteristics}
\end{table*}

\end{document}